\documentclass[11pt,onecolumn,letterpaper]{article}
\usepackage[letterpaper,left=0.70in,right=0.70in,top=0.78in,bottom=0.78in,headheight=24pt,headsep=16pt,footskip=25pt]{geometry}
\usepackage[T1]{fontenc}
\usepackage[type1,lining]{ebgaramond}
\usepackage{fix-cm}
\usepackage{graphicx,amsmath,amssymb,booktabs}
\usepackage{ebgaramond-maths}
\DeclareSymbolFont{fallbackletters}{OML}{cmm}{m}{it}
\SetSymbolFont{fallbackletters}{bold}{OML}{cmm}{b}{it}
\DeclareMathSymbol{\epsilon}{\mathord}{fallbackletters}{"0F}
\usepackage[dvipsnames]{xcolor}
\usepackage[numbers,sort&compress]{natbib}
\usepackage{caption,subcaption,etoolbox}
\usepackage{fancyhdr,titlesec,placeins}
\usepackage[most]{tcolorbox}
\usepackage[shortlabels,inline]{enumitem}
\usepackage{xspace,fontawesome5}
\DeclareRobustCommand{\weathercraft}{\mbox{\normalfont\scshape WeatherCraft}}
\usepackage{pifont} % Filled circled contribution numbers.
\usepackage{tabularx,colortbl}
\usepackage{algorithm}
\usepackage{algpseudocode}
\usepackage{placeins} % Keep the paper's pending figures before its references.

\definecolor{reportink}{HTML}{252335}
\definecolor{reportaccent}{HTML}{655080}
\definecolor{reportrule}{HTML}{DCD4E7}
\definecolor{reportfront}{HTML}{F6F8FB}
\usepackage[breaklinks,colorlinks,linkcolor=blue!55!black,citecolor=blue!55!black,urlcolor=blue!60!black]{hyperref}
\usepackage[capitalize,nameinlink]{cleveref}
\crefname{figure}{Fig.}{Figs.}
\Crefname{figure}{Figure}{Figures}
\hypersetup{pdftitle={Reinforcement-Aligned Probabilistic Weather Forecasting with Discrete Tokens},pdfauthor={Jinhan Li, Zhisheng Chen, Yuxuan Li, Yuan Gao, Fan Xu, Hao Wu, Jinlong Du, Kun Wang}}
\titleformat{\section}{\Large\bfseries\color{reportink}}{\thesection}{0.65em}{}
\titleformat{\subsection}{\large\bfseries\color{reportink}}{\thesubsection}{0.65em}{}
\titlespacing*{\section}{0pt}{16pt plus 3pt minus 2pt}{7pt}
\titlespacing*{\subsection}{0pt}{11pt plus 2pt minus 1pt}{5pt}
\renewcommand{\headrulewidth}{0.4pt}
\renewcommand{\headrule}{\hbox to\headwidth{\color{reportrule}\leaders\hrule height\headrulewidth\hfill}}
\fancypagestyle{firstpage}{\fancyhf{}\fancyfoot[C]{\small\thepage}\renewcommand{\headrulewidth}{0pt}}
\makeatletter
\newenvironment{reportfrontmatter}{%
 \thispagestyle{firstpage}%
 \begin{tcolorbox}[enhanced,colback=reportfront,frame hidden,boxrule=0pt,
 arc=11pt,left=16pt,right=16pt,top=16pt,bottom=15pt,
 boxsep=0pt,before skip=0pt,after skip=15pt]
 {\raggedright\fontsize{24}{27}\selectfont\bfseries\@title\par}
 \vspace{9pt}
 {\fontsize{11.8}{14.2}\selectfont\bfseries\reportauthors\par}
 \vspace{8pt}
 {\fontsize{10.8}{13.3}\selectfont\reportaffiliations\par}
 \vspace{3pt}
 {\small\reportauthornotes\par}
}{\end{tcolorbox}}
\renewenvironment{abstract}{\par\vspace{11pt}\begingroup\normalsize\setlength{\parindent}{0pt}\ignorespaces}{\par\endgroup}
\makeatother
\usepackage{microtype}
\makeatletter
\providecommand{\theHALG@line}{}
\renewcommand{\theHALG@line}{\thealgorithm.\arabic{ALG@line}}
\makeatother

\usepackage{needspace}
\pretocmd{\section}{\Needspace{6\baselineskip}}{}{}

\usepackage{wrapfig}

\newcommand{\reportshorttitle}{WeatherCraft}
\newcommand{\reporttype}{Technical Report}
\title{Reinforcement-Aligned Probabilistic Weather Forecasting with Discrete Tokens}
\author{Jinhan Li \and Zhisheng Chen \and Yuxuan Li \and Yuan Gao \and Fan Xu \and Hao Wu \and Jinlong Du \and Kun Wang}
\newcommand{\reportauthors}{Jinhan Li$^{1}$ \quad Zhisheng Chen$^{2,\dagger}$ \quad Yuxuan Li$^{2}$ \quad Yuan Gao$^{3}$\\[3pt]
Fan Xu$^{1}$ \quad Hao Wu$^{4,\dagger}$ \quad Jinlong Du$^{5}$ \quad Kun Wang$^{2}$}
\newcommand{\reportaffiliations}{$^{1}$University of Science and Technology of China\\
$^{2}$Nanyang Technological University \quad $^{3}$Tsinghua University\\
$^{4}$Carnegie Mellon University \quad $^{5}$Tongji University}
\newcommand{\reportauthornotes}{$^{\dagger}$Corresponding authors: Zhisheng Chen and Hao Wu\\
\href{mailto:zhisheng.researcher@gmail.com}{zhisheng.researcher@gmail.com} \quad \href{mailto:hw5@andrew.cmu.edu}{hw5@andrew.cmu.edu}}

\begin{document}
\begin{reportfrontmatter}
\begin{abstract}
Global medium-range weather forecasting must not only generate ensembles efficiently to characterize uncertainty in future states, but also serve downstream applications such as extreme-event warnings and cross-sphere coupled simulations. We introduce \weathercraft{}, a framework that unifies probabilistic generation and reinforcement learning alignment through discrete weather tokens, bringing the GPT-style pretraining--post-training paradigm to global medium-range weather forecasting. A multiscale tokenizer represents atmospheric states as a coarse-to-fine token hierarchy, while a Transformer models causal dependencies across time and scale and predicts tokens within each scale in parallel. Direct sampling generates diverse ensembles, and exact token log-probabilities enable Group Relative Policy Optimization (GRPO) to maximize downstream task rewards without differentiating through evaluators or coupled models. On the ERA5 benchmark, \weathercraft{}'s eight-member ensemble mean reduces day-$15$ normalized RMSE by $17.6\%$ relative to the strongest evaluated deterministic baseline and, with KV caching, achieves an approximately $18\times$ inference speedup over GenCast. Direct fair-CRPS adaptation improves ensemble probabilistic skill, while reward-based post-training improves forecasts of persistent land temperature extremes and coupled ocean heat extremes. This work establishes discrete autoregressive modeling as a general foundation for scalable, task-alignable Earth system foundation models.
\end{abstract}

\par\vspace{8pt}
{\small\noindent\faGithub\enspace\textbf{Code}\quad
 \url{https://github.com/Jinhan-Li/WeatherCraft}\par}
\end{reportfrontmatter}
\raggedbottom
\section{Introduction}
\label{sec:intro}

Global medium-range weather forecasting faces two complementary demands: the chaotic nature of atmospheric dynamics calls for efficient ensemble generation to characterize uncertainty in future states~\cite{lorenz1963chaos,leutbecher2008ensemble}, while downstream applications such as extreme-event warnings and cross-sphere coupled simulations require adaptation to specific task objectives. Data-driven models substantially advance medium-range accuracy~\cite{rasp2024weatherbench2}. Unifying efficient ensemble generation and task-driven policy optimization within a single generative framework is therefore a fundamental challenge for next-generation weather forecasting.

\begin{wrapfigure}{r}{0.40\textwidth}
    \vspace{-8pt}
    \captionsetup{font=footnotesize}
    
    \centering
    \includegraphics[width=\linewidth]{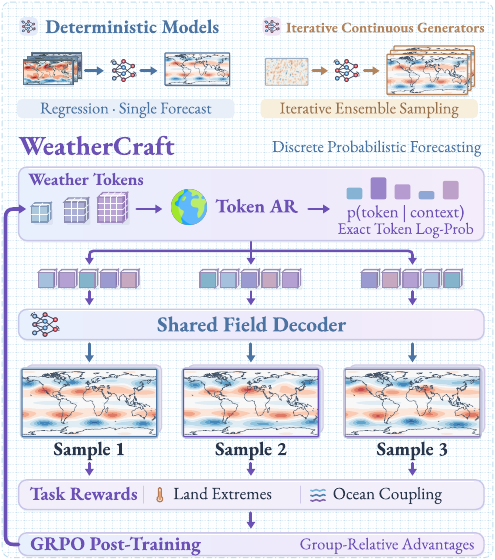}
    \caption{\textbf{Weather forecasting paradigms.} Deterministic models output one trajectory (top left), while iterative continuous generators repeatedly denoise samples (top right). \weathercraft{} (bottom) samples discrete weather tokens and aligns its generative policy with downstream task rewards via GRPO.}
    \label{fig:paradigms}
\vspace{-6pt}
\end{wrapfigure}

As shown in Fig.~\ref{fig:paradigms} (top left), deterministic models, including Pangu-Weather~\cite{bi2023pangu}, GraphCast~\cite{lam2023graphcast}, FengWu~\cite{chen2025fengwu}, FuXi~\cite{chen2023fuxi}, and OneForecast~\cite{gao2025oneforecast}, predict future states through continuous-field regression. These models deliver strong short-range point forecasts, but standard inference outputs a single trajectory rather than an explicit distribution over future states. Continuous generative models, such as diffusion~\cite{ho2020ddpm} and flow matching~\cite{lipman2023flowmatching} (e.g., GenCast~\cite{price2025gencast}), instead construct ensembles by transforming noise into forecasts (Fig.~\ref{fig:paradigms}, top right). Iterative samplers require repeated network evaluations, increasing inference cost with ensemble size and forecast horizon; one-step flow models such as Tyche~\cite{xu2026tyche} reduce this cost, although our daily-resolution evaluation reveals long-range skill degradation and a year-long rollout cold bias (Fig.~\ref{fig:annual}). Moreover, continuous sampling does not directly expose step-wise discrete likelihoods for likelihood-ratio policy optimization.

\Needspace{0.55\textheight}

This limitation is particularly acute in Earth system applications. Evaluating persistent land temperature extremes involves quantile thresholds and multi-day persistence rules. Marine heatwave forecasting tracks extremes in sea surface temperature~\cite{shu2025advancedmhw}, which can be predicted by forcing an ocean model with atmospheric forecasts~\cite{wu2025tritoncast}. Threshold-based event scores are non-differentiable, and backpropagating through a coupled simulator incurs substantial overhead. Although BeamVQ~\cite{wang2025beamvq} incorporates task feedback via quantized candidate search and self-training, making direct policy adaptation to extreme-event and coupled-ocean rewards an intrinsic capability of a global multivariate weather generator remains challenging. Autoregressive modeling over discrete tokens~\cite{oord2017vqvae,esser2021vqgan} provides a natural interface: in the GPT pretraining--post-training paradigm~\cite{radford2019gpt2,ouyang2022instructgpt,gallici2025vargrpo}, categorical heads yield exact conditional token log-probabilities, allowing scalar rewards to guide policy updates without differentiating through the evaluator or coupled model~\cite{williams1992reinforce}. Bringing this paradigm to global weather forecasting requires representing planetary circulation and localized gradients across dozens of coupled channels while decoding tens of thousands of tokens in parallel.

We introduce \weathercraft{} (Fig.~\ref{fig:paradigms}, bottom), unifying probabilistic generation and reinforcement learning alignment via discrete weather tokens. \weathercraft{} uses a multi-branch, multiscale spatial residual tokenizer based on finite scalar quantization (FSQ)~\cite{mentzer2024fsq} to encode atmospheric states into a high-capacity coarse-to-fine token hierarchy without learning a vector codebook. Following next-scale prediction~\cite{tian2024var}, a conditional Transformer takes the preceding two days of latents as context and generates the next-day token hierarchy causally across time and scale while predicting all spatial tokens within each scale in parallel; multi-day rollouts feed generated representations back directly in latent space without repeated field decoding and re-encoding. This design supports fast ensemble sampling and yields exact token log-probabilities for every trajectory. Building on this interface, we treat the pretrained dynamics model as a generative policy and perform post-training with Group Relative Policy Optimization (GRPO)~\cite{shao2024grpo} over branching rollout trees, directly optimizing land extreme-event metrics or coupled responses of a frozen ocean model without differentiating through downstream evaluators.

We systematically evaluate \weathercraft{} on a global $69$-channel daily-mean ERA5 benchmark~\cite{hersbach2020era5}. Its eight-member ensemble mean closely matches the strongest deterministic baseline at day $1$ and achieves the lowest overall normalized RMSE (nRMSE) and highest anomaly correlation coefficient (ACC) among the evaluated deterministic baselines across all reported $3$--$15$-day leads, reducing day-$15$ nRMSE by $17.6\%$ relative to the strongest deterministic baseline; with KV caching, eight-member $15$-day inference is $17.9\times$ faster than GenCast. Two-stage fair-CRPS adaptation further improves probabilistic scores across leads, narrowing the day-$15$ gap to GenCast to within $1\%$. Tree GRPO raises mean land-extreme CSI by $0.00554$ and closes $7.38\%$ of the SST exceedance CSI gap to ERA5 forcing. In missing-input tests, replacing a surface variable with climatology has little effect on its forecast RMSE. Our main contributions are as follows:

\smallskip
\noindent\ding{182}~\textbf{\textit{An efficient discrete probabilistic weather forecasting framework.}} We propose \weathercraft{}, combining multi-branch, multiscale residual FSQ representations with a causal time--scale Transformer for accurate, efficient ensemble generation through within-scale parallel decoding and latent-space rollout across forecast days.

\smallskip
\noindent\ding{183}~\textbf{\textit{Reinforcement learning alignment without downstream differentiation.}} Leveraging exact token log-probabilities and branching rollout-tree GRPO, we establish a unified pretraining--post-training framework that optimizes non-differentiable extreme-event metrics and coupled ocean forecasting skill without differentiating through evaluators.

\smallskip
\noindent\ding{184}~\textbf{\textit{Systematic validation across global forecasting and land--ocean tasks.}} We comprehensively evaluate medium-range ensemble forecasting, extended rollouts, missing-variable recovery, and land--ocean extreme-event tasks, demonstrating the effectiveness of discrete generation and reward alignment for Earth system modeling.

\section{Related Work}
\label{sec:related}

\paragraph{Data-driven weather forecasting.}
Deterministic global forecasters regress atmospheric fields to produce a single trajectory~\cite{pathak2022fourcastnet,bi2023pangu,lam2023graphcast,chen2025fengwu,chen2023fuxi,gao2025oneforecast}. FuXi-ENS learns perturbations for ensemble forecasting~\cite{zhong2025fuxiens}, and AIFS-CRPS optimizes the continuous ranked probability score~\cite{lang2026aifscrps}. FGN and Mosaic generate ensemble members through learned functional perturbations~\cite{alet2025fgn,zhdanov2026mosaic}. Mosaic processes the native weather grid with block-sparse attention to preserve spatial detail. WeatherNext Cyclones extends FGN to jointly predict global weather and cyclone tracks, intensity, and size~\cite{alet2026cyclones}. GenCast samples atmospheric fields through conditional diffusion~\cite{price2025gencast}, and ArchesWeatherGen samples deterministic forecast residuals through flow matching~\cite{couairon2026archesweathergen}. Iterative sampling costs accumulate across ensemble members and forecast days. Tyche uses one flow-model evaluation per step~\cite{xu2026tyche}, and Swift pairs single-step consistency sampling with autoregressive CRPS fine-tuning~\cite{stock2026swift}. OmniCast uses masked generation and per-token diffusion in a continuous latent space~\cite{nguyen2025omnicast}. \textit{\weathercraft{} samples discrete weather tokens. The same probabilities used to generate each member yield its trajectory log-probability for policy updates from downstream rewards.}

\paragraph{Discrete tokens and autoregressive visual generation.}
Discrete representation learning maps continuous signals to symbols in a finite vocabulary. VQ-VAE~\cite{oord2017vqvae} and VQGAN~\cite{esser2021vqgan} learn vector codebooks; RQ-VAE~\cite{lee2022rq} progressively refines representations through residual quantization. FSQ~\cite{mentzer2024fsq} replaces codebook learning with bounded scalar quantization, while MAGVIT-v2~\cite{yu2024lfq} supports large vocabularies through lookup-free quantization and vocabulary factorization. Token-based visual generators adopt different decoding orders: LlamaGen~\cite{sun2024llamagen} predicts tokens in raster order, MaskGIT~\cite{chang2022maskgit} uses iterative masked parallel decoding, and VAR~\cite{tian2024var} predicts entire token maps from coarse to fine. In Earth science, GPTCast~\cite{franch2025gptcast} combines VQGAN and GPT to generate probabilistic radar precipitation nowcasts through categorical sampling, while ClimateAR~\cite{yu2026climatear} applies multiscale autoregressive modeling to climate forecasting. Global multivariate medium-range forecasting demands high-fidelity representations of coupled physical fields and efficient generation of large daily token grids. \textit{\weathercraft{} meets these demands via a spatial residual hierarchy of multiple FSQ groups and causal time--scale modeling with parallel within-scale decoding.}

\paragraph{Reinforcement learning post-training for generative models.}
RLHF aligns language models with human preferences through learned rewards~\cite{ouyang2022instructgpt}, while GRPO replaces a learned value network with group-relative rewards~\cite{shao2024grpo}. TreeRL~\cite{hou2025treerl}, TreePO~\cite{li2026treepo}, and Tree-GRPO~\cite{ji2025treegrpo} use shared-prefix trees to assign credit from outcome rewards. For visual generation, DDPO~\cite{black2024ddpo} formulates denoising as sequential decision-making, and Flow-GRPO~\cite{liu2025flowgrpo} enables policy-gradient optimization of flow models through stochastic sampling. Diffusion-DPO~\cite{wallace2024diffusiondpo} instead optimizes preferences using a variational likelihood objective. SimpleAR~\cite{wang2025simplear}, AR-GRPO~\cite{yuan2025argrpo}, and T2I-R1~\cite{jiang2025t2ir1} use autoregressive token log-probabilities and GRPO to improve image quality and text alignment. Task feedback also guides physical spatiotemporal forecasting: BeamVQ~\cite{wang2025beamvq} selects quantized candidates with non-differentiable metrics such as CSI for self-training; SynCast~\cite{xu2025syncast} applies sequential preference optimization to diffusion nowcasting to balance CSI and false alarms; and FAKER-Air~\cite{kang2026fakerair} uses class-wise rewards for GRPO-based regional air-quality forecasting. \textit{\weathercraft{} targets global multivariate medium-range forecasting, using exact discrete-token likelihoods for policy-gradient optimization with rewards from non-differentiable event metrics or frozen ocean models. Branching rollout trees assign credit across forecast time.}

% LaTeX subfile: load amsmath and amssymb in the parent document.
% Bibliography and document environment are managed by the parent.

\begin{figure*}[t]
    \centering
    \includegraphics[width=\textwidth]{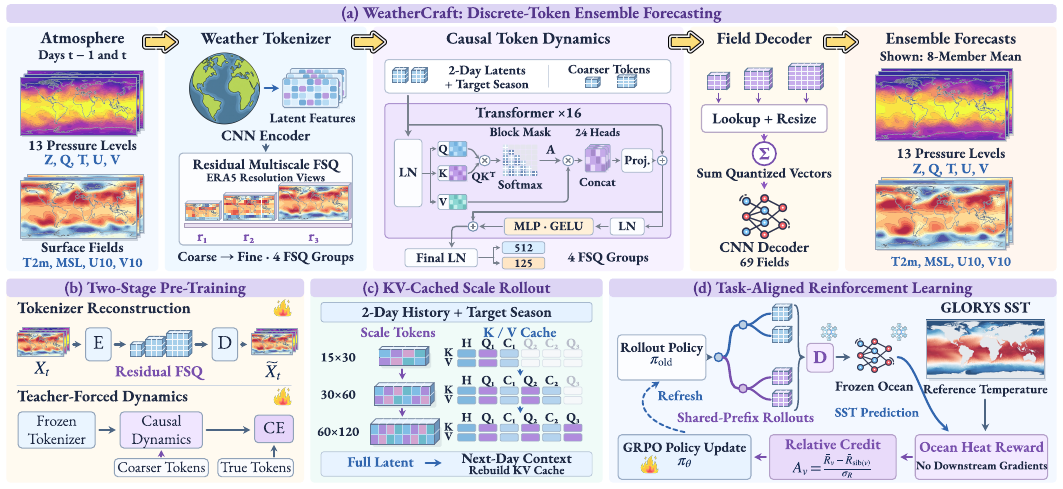}
    \caption{\textbf{Architecture and learning framework of \weathercraft{}.} (a) Multiscale FSQ tokenization, block-causal dynamics, and field decoding; (b) pretraining; (c) KV-cached latent rollout; and (d) rollout-tree GRPO. Flame/snowflake icons mark trainable/frozen modules.}
    \label{fig:architecture}
\end{figure*}

\section{Methodology}
\label{sec:methodology}

\weathercraft{} formulates global medium-range weather forecasting as discrete multiscale generation followed by reward-driven policy alignment (Fig.~\ref{fig:architecture}). Rather than regressing deterministic fields or iteratively denoising continuous states, our framework represents daily atmospheric states as coarse-to-fine pyramids of discrete tokens (Sec.~\ref{sec:representation}), models their temporal and cross-scale transitions through a block-causal Transformer in latent space (Sec.~\ref{sec:dynamics}), and directly aligns the resulting generative policy with downstream task rewards through exact token log-likelihoods (Secs.~\ref{sec:search} and~\ref{sec:optimization}).

\subsection{Problem Formulation}
\label{sec:problem}
Let $X_t\in\mathbb R^{C\times H\times W}$ denote the standardized daily atmospheric state at day $t$, comprising $C$ physical channels on an $H\times W$ global grid. Following standard two-step Markov conditioning in global weather forecasting~\cite{lam2023graphcast,chen2025fengwu,price2025gencast}, we condition the next-day transition distribution $\mathcal F_\Theta$ on $(X_{t-1},X_t)$ and the target-day annual phase $c_{t+1}\in\mathbb R^2$:
\begin{equation}
\widehat X_{t+1}\sim\mathcal F_\Theta(X_{t-1},X_t,c_{t+1}).
\label{eq:method-1}
\end{equation}
Repeated ancestral sampling and multi-day autoregressive rollout yield a diverse ensemble of forecast trajectories.

\subsection{Multiscale Residual FSQ Tokenization}
\label{sec:representation}
Global multivariate weather fields combine large-scale spatial coherence with localized synoptic gradients and strong cross-variable coupling. To represent planetary circulation patterns and fine-scale structure with compact per-group vocabularies, we combine a spatial residual hierarchy~\cite{lee2022rq,tian2024var,yu2026climatear} with multi-group FSQ~\cite{mentzer2024fsq}.

A convolutional encoder $\mathcal E$ with circular longitude padding and reflecting latitude padding maps $X$ to a continuous latent representation $f=\mathcal E(X)$, which is partitioned along the channel axis into $G$ parallel groups of dimension $D$. This multi-group decomposition expands representational capacity multiplicatively across channel subspaces while keeping each per-group vocabulary compact. Across $S$ coarse-to-fine spatial scales starting from $\widehat f_g^0=0$, each group $g\in\{1,\ldots,G\}$ recursively quantizes the residual left by coarser scales via spatial downsampling $\mathrm{down}(\cdot,s)$, scalar quantization $\mathcal Q(\cdot)$, and upsampling $\mathrm{up}(\cdot,S)$:
\begin{equation}
\begin{gathered}
r_g^s=\mathcal Q\!\left(\mathrm{down}(f_g-\widehat f_g^{s-1},s)\right),\\
\widehat f_g^s=\widehat f_g^{s-1}+\mathrm{up}\!\left(\mathrm{lookup}(r_g^s),S\right).
\end{gathered}
\label{eq:method-2}
\end{equation}
Stacking the group indices $r_g^s$ gives the scale-$s$ token map $r^s$. Coarse tokens can encode planetary-scale circulation patterns such as Rossby waves, and finer tokens can progressively resolve localized synoptic-scale residuals across spatial scales.

For a local residual vector $u\in\mathbb R^D$ with axis levels $(L_1,\ldots,L_D)$ and bounding function $b_d(\cdot)$, FSQ rounds each coordinate via a straight-through estimator ($\operatorname{round}_{\mathrm{STE}}$)~\cite{bengio2013ste,oord2017vqvae} into a discrete vocabulary of size $V=\prod_{d=1}^D L_d$:
\begin{equation}
\begin{gathered}
\widehat u_d=\frac{\operatorname{round}_{\mathrm{STE}}(b_d(u_d))}{\lfloor L_d/2\rfloor},
\quad d=1,\ldots,D,\\
\mathcal Q(u)=\operatorname{index}(\widehat u),\qquad
\mathrm{lookup}(\mathcal Q(u))=\widehat u.
\end{gathered}
\label{eq:method-3}
\end{equation}
Unlike vector quantization with a learned codebook, this fixed Cartesian grid requires neither learned codebook vectors nor auxiliary commitment losses. Summing the upsampled corrections across all $S$ scales and concatenating the $G$ groups yields the daily latent $z=\widehat f^S$, which the convolutional decoder $\mathcal D$ maps to $\widetilde X$ (Sec.~\ref{app:a-tokenizer}):
\begin{equation}
z=\sum_{s=1}^S\mathrm{up}\!\left(\mathrm{lookup}(r^s),S\right),\qquad
\widetilde X=\mathcal D(z).
\label{eq:method-4}
\end{equation}

\subsection{Block-Causal Autoregressive Dynamics}
\label{sec:dynamics}
Let $r_t=(r_t^1,\ldots,r_t^S)=\mathcal T(X_t)$ denote the daily token pyramid with reconstructed latent $z_t=\widehat f_t^S$ and coarser-scale prefix $r_t^{<s}=(r_t^1,\ldots,r_t^{s-1})$. Given the two-day latent context $h_t=(z_{t-1},z_t,c_{t+1})$, the conditional policy $\pi_\theta$ factorizes single-day generation from coarse to fine across scales, and factorizes a $T$-day rollout autoregressively across days:
\begin{equation}
\begin{gathered}
\pi_\theta(r_{t+1}\mid h_t)=\prod_{s=1}^S\pi_\theta(r_{t+1}^s\mid r_{t+1}^{<s},h_t),\\
\pi_\theta(r_{t+1:t+T}\mid h_t)=\prod_{\ell=1}^T\pi_\theta(r_{t+\ell}\mid h_{t+\ell-1}).
\end{gathered}
\label{eq:method-7}
\end{equation}
During multi-day rollouts, each ensemble member updates its conditioning context directly in latent space via $h_{t+\ell}=(z_{t+\ell-1},z_{t+\ell},c_{t+\ell+1})$, avoiding repeated physical-field decoding and re-encoding across rollout steps (Fig.~\ref{fig:architecture}(c)).

To reconcile cross-scale causality with parallel spatial decoding, the Transformer processes a block-structured sequence of the bidirectional two-day history $\mathsf H$ and interleaved target-day query and content blocks $(\mathsf U^s,\mathsf V^s)$:
\begin{equation}
Y_0 = [\mathsf H;\,\mathsf U^1;\,\mathsf V^1;\,\ldots;\,\mathsf U^S],
\qquad M_{ij}=\mathbf 1[\beta(i)\ge\beta(j)].
\label{eq:method-5}
\end{equation}
At scale $s$, the query block $\mathsf U^s$ embeds $z_t$ and the target latent reconstructed through scale $s-1$. The content block $\mathsf V^s$ embeds the same $z_t$ and the target latent updated with scale-$s$ tokens. For block indices $\beta(i)$ and $\beta(j)$, $M_{ij}=1$ lets position $i$ attend to $j$ in the same or an earlier block. At inference, key--value projections from completed blocks are cached and reused across all subsequent spatial scales (detailed in Sec.~\ref{app:a-dynamics}).

We factorize each token $q$ into two categorical factors $(q^A,q^B)$~\cite{yu2024lfq}, giving $512\times125=64{,}000$ codes per group. At cell $u\in\Omega_s$, two independent linear heads predict each group's factors from the same query representation $y_u^s$. Summing their log-probabilities over cells and groups gives the exact scale-$s$ log-likelihood:
\begin{equation}
\begin{aligned}
\log\pi_\theta(r_{t+1}^s\mid r_{t+1}^{<s},h_t)
=\sum_{u\in\Omega_s}\sum_{g=1}^G\Big[
 &\log p_\theta^{(A)}(q_{u,g}^A\mid y_u^s)\\[-2pt]
 &+\log p_\theta^{(B)}(q_{u,g}^B\mid y_u^s)\Big].
\end{aligned}
\label{eq:method-6}
\end{equation}

\subsection{Rollout-Tree Reinforcement Learning}
\label{sec:search}
Downstream Earth system tasks often evaluate forecasts through non-differentiable event criteria or coupled physical simulators. Because Eq.~\eqref{eq:method-6} exposes exact token log-probabilities, we align $\pi_\theta$ directly with downstream task rewards via reinforcement learning without differentiating through the downstream evaluator (Fig.~\ref{fig:architecture}(d)).

Given a frozen policy $\pi_{\mathrm{old}}$ and context $h_t$, a rollout operator $\mathcal S$ samples a batch $\mathcal B=\mathcal S(\pi_{\mathrm{old}},h_t;K)$ of $K$ decoded trajectories $\{\tau_i\}_{i=1}^K$ scored by task reward $R_i=R(\tau_i)$:
\begin{equation}
\mathcal B=\mathcal S(\pi_{\mathrm{old}},h_t;K),\qquad
R_i=R(\tau_i),\quad i=1,\ldots,K.
\label{eq:method-8}
\end{equation}
For land extremes, $R$ measures the critical success index (CSI) of persistent warm and cold temperature events; for ocean coupling, predicted atmospheric fields force a frozen ocean circulation model~\cite{wu2025tritoncast} and $R$ scores the resulting coupled sea-surface temperature extremes (Sec.~\ref{app:c-downstream}).

Assigning a single trajectory return to all forecast days confounds early synoptic decisions with late-stage divergence. Our rollout operator $\mathcal S$ instead constructs a shared-prefix branching tree that bifurcates at intermediate forecast leads, where children share their parent's latent prefix and sample distinct future continuations. For a newly generated tree segment $v$ with descendant leaf set $\mathcal L(v)$, sibling branch $\mathrm{sib}(v)$, and leaf reward standard deviation $\sigma_R$, we adapt GRPO~\cite{shao2024grpo} into a sibling-relative advantage:
\begin{equation}
\bar R_v=\frac{1}{|\mathcal L(v)|}\sum_{i\in\mathcal L(v)}R_i,\qquad
A_v=\frac{\bar R_v-\bar R_{\mathrm{sib}(v)}}{\sigma_R}.
\label{eq:method-9}
\end{equation}
Every token occurrence $e\in v$ newly sampled in segment $v$ inherits $A_e=A_v$, while shared prefix segments are stored and credited once (Sec.~\ref{app:c-grpo}). Comparing sibling continuations under a shared prefix provides a relative learning signal for each branching horizon, rather than assigning one trajectory return uniformly across forecast days.

\subsection{Optimization Objectives}
\label{sec:optimization}
\paragraph{Stages 1 and 2: Tokenizer and dynamics pretraining.}
Let $\|\cdot\|_w^2$ denote the latitude-weighted mean squared error in standardized space and $N$ be the daily composite token count. Stage 1 trains the autoencoder parameters $\phi=(\mathcal E,\mathcal D)$ by minimizing reconstruction error $\mathcal L_{\mathrm{tok}}(\phi)$; Stage 2 freezes $\mathcal T$ and trains the dynamics policy $\theta$ via teacher-forced next-day negative log-likelihood $\mathcal L_{\mathrm{dyn}}(\theta)$ (Fig.~\ref{fig:architecture}(b)):
\begin{equation}
\begin{aligned}
\mathcal L_{\mathrm{tok}}(\phi) &= \mathbb E_X\!\left[\|X-\widetilde X\|_w^2\right],\\
\mathcal L_{\mathrm{dyn}}(\theta) &= -\frac1N\mathbb E\!\left[\sum_{s=1}^S\log\pi_\theta(r_{t+1}^s\mid r_{t+1}^{<s},h_t)\right].
\end{aligned}
\label{eq:method-10}
\end{equation}

\paragraph{Stage 3: Clipped GRPO policy optimization.}
We initialize $\pi_\theta$ from Stage 2 and freeze the reference policy $\pi_{\mathrm{ref}}$. For each unique token occurrence $e\in\mathcal B$, $a_e$ is the sampled token and $h_e$ contains its two-day latent history, target-day annual phase, and coarser target tokens. Each occurrence has probability ratio $\rho_e=\pi_\theta(a_e\mid h_e)/\pi_{\mathrm{old}}(a_e\mid h_e)$, clipped ratio $\bar\rho_e=\operatorname{clip}(\rho_e,1-\epsilon,1+\epsilon)$, normalized scale-balanced weight $w_e$ (aggregating coarse, medium, and fine scales in the ratio $1{:}2{:}4$), and non-negative reference KL penalty $k_e=e^{\delta_e}-\delta_e-1$ with $\delta_e=\log(\pi_{\mathrm{ref}}(a_e\mid h_e)/\pi_\theta(a_e\mid h_e))$. With clipping radius $\epsilon$ and reference-policy KL weight $\beta$, we maximize the clipped surrogate objective:
\begin{equation}
\mathcal J(\theta)=\sum_{e\in\mathcal B}w_e
\left[\min\{\rho_e A_e,\bar\rho_e A_e\}-\beta k_e\right].
\label{eq:method-12}
\end{equation}
Rollout probabilities, advantages, and weights remain fixed within each update step. At inference, ensemble members are sampled directly from the aligned policy $\pi_\theta$ without reward-based selection or post-hoc filtering.

\paragraph{Direct fair-CRPS adaptation.}
Separately from GRPO, we initialize a score-based branch from the pretrained policy and optimize fair-CRPS via a $1$-to-$2$-day rollout curriculum ($T\in\{1,2\}$; Sec.~\ref{app:c-crps}). Using Gumbel-Softmax straight-through samples $\widetilde a=\operatorname{onehot}(\arg\max(\lambda+g))+p_\tau-\operatorname{sg}(p_\tau)$ from factor logits $\lambda$ to query fixed FSQ vectors through frozen $\mathcal D$, we minimize two-member terminal fair-CRPS~\cite{ferro2014fairscores,xu2026tyche} under weighted $L_1$ distance $d_w$:
\begin{equation}
\begin{aligned}
\mathcal L_{\mathrm{CRPS}}^{(T)}(\theta)
=\tfrac12\Big[
 &d_w(\widehat X_{t+T}^{(1)},X_{t+T})+d_w(\widehat X_{t+T}^{(2)},X_{t+T})\\[-2pt]
 &-d_w(\widehat X_{t+T}^{(1)},\widehat X_{t+T}^{(2)})\Big].
\end{aligned}
\label{eq:method-13}
\end{equation}
Hard forward samples preserve discrete rollout while biased straight-through gradients propagate through the frozen decoder inputs across all generated scales and forecast days.

\section{Experiments}
\label{sec:experiments}

% Verified source values preserved; styled with EB Garamond and ranked cell highlights.
\begingroup
\definecolor{resultink}{HTML}{292C42}
\definecolor{resultpurple}{HTML}{544178}
\definecolor{resultheader}{HTML}{E9E8F2}
\definecolor{resultstripe}{HTML}{F6F7FA}
\definecolor{resultours}{HTML}{EEE7F7}
\definecolor{rankfirst}{HTML}{E4D9F1}
\definecolor{ranksecond}{HTML}{F7E7D6}
\definecolor{rankthird}{HTML}{E0EDF4}
\newcommand{\rankone}[1]{\cellcolor{rankfirst}\textbf{#1}}
\newcommand{\ranktwo}[1]{\cellcolor{ranksecond}#1}
\newcommand{\rankthree}[1]{\cellcolor{rankthird}#1}
\newcommand{\resultcategory}[1]{%
  \addlinespace[1.5pt]
  \multicolumn{11}{@{}l}{\rule{0pt}{10.5pt}\textcolor{resultpurple}{\bfseries\itshape #1}}\\[0.5pt]}
\newcommand{\dailymetrics}{%
  {\fontsize{8.5}{9.6}\selectfont\mbox{nRMSE$\downarrow$}} &
  {\fontsize{8.5}{9.6}\selectfont\mbox{ACC$\uparrow$}}}

\begin{table*}[t]
\centering
\caption{\textbf{Quantitative comparison of global daily weather forecasting} across $1{,}081$ test initializations (2020--2022, $69$ ERA5 channels) in latitude-weighted nRMSE$\downarrow$ and ACC$\uparrow$. \weathercraft{} uses an $8$-member mean; \colorbox{rankfirst}{\textbf{1st}}, \colorbox{ranksecond}{2nd}, and \colorbox{rankthird}{3rd} mark the top three scores.}
\label{tab:daily}
\vspace{2pt}
\fontfamily{EBGaramond-LF}\fontsize{9.2}{10.8}\selectfont
\color{resultink}
\setlength{\tabcolsep}{1.5pt}
\renewcommand{\arraystretch}{1.05}
\setlength{\aboverulesep}{1.8pt}
\setlength{\belowrulesep}{1.8pt}
\begin{tabularx}{\textwidth}{@{}p{0.19\textwidth}*{10}{>{\centering\arraybackslash}X}@{}}
\toprule[0.95pt]
\rowcolor{resultheader}
\rule{0pt}{10.5pt}\textbf{Method}
 & \multicolumn{2}{c}{\textbf{1-Day}}
 & \multicolumn{2}{c}{\textbf{3-Day}}
 & \multicolumn{2}{c}{\textbf{5-Day}}
 & \multicolumn{2}{c}{\textbf{10-Day}}
 & \multicolumn{2}{c@{}}{\textbf{15-Day}}\\[-0.5pt]
\cmidrule(lr){2-3}\cmidrule(lr){4-5}\cmidrule(lr){6-7}\cmidrule(lr){8-9}\cmidrule(l){10-11}
 & \dailymetrics & \dailymetrics & \dailymetrics & \dailymetrics & \dailymetrics\\[-0.5pt]
\midrule[0.55pt]
\resultcategory{General Spatiotemporal Prediction Models}
\rowcolor{resultstripe} ConvLSTM~\cite{shi2015convlstm} & 0.3410 & 0.8302 & 0.7440 & 0.3598 & 0.9497 & 0.1066 & 1.4206 & -0.0205 & 1.5766 & -0.0290 \\
SimVP~\cite{gao2022simvp} & 0.3032 & 0.8231 & 0.4521 & 0.6529 & 0.5877 & 0.4541 & 0.7518 & 0.1566 & 0.8169 & 0.0666 \\
\resultcategory{Data-Driven Weather Forecasting Models}
\rowcolor{resultstripe} SFNO~\cite{bonev2023sfno} & 0.2751 & 0.8895 & 0.4508 & 0.6762 & 0.5587 & 0.4829 & 0.6829 & 0.1872 & \ranktwo{0.7245} & 0.0940 \\
Pangu-Weather~\cite{bi2023pangu} & 0.2070 & 0.9370 & 0.4283 & 0.7331 & 0.5838 & 0.4870 & 0.7313 & 0.1648 & 0.7715 & 0.0850 \\
\rowcolor{resultstripe} OneForecast~\cite{gao2025oneforecast} & \rankone{0.1444} & \rankone{0.9703} & \ranktwo{0.2618} & \ranktwo{0.9041} & \ranktwo{0.3920} & \ranktwo{0.7866} & \ranktwo{0.6594} & \ranktwo{0.3944} & 0.7680 & \ranktwo{0.1866} \\
FuXi~\cite{chen2023fuxi} & 0.1794 & 0.9532 & 0.3457 & 0.8346 & 0.4980 & 0.6510 & 0.7091 & 0.2758 & 0.7716 & 0.1402 \\
\rowcolor{resultstripe} FengWu~\cite{chen2025fengwu} & 0.1943 & 0.9435 & 0.3366 & 0.8229 & 0.4684 & 0.6551 & 0.6694 & 0.2886 & \rankthree{0.7397} & 0.1432 \\
GraphCast~\cite{lam2023graphcast} & \rankthree{0.1546} & \rankthree{0.9653} & \rankthree{0.2776} & \rankthree{0.8894} & \rankthree{0.4081} & \rankthree{0.7611} & \rankthree{0.6608} & \rankthree{0.3616} & 0.7569 & \rankthree{0.1671} \\
\rowcolor{resultstripe} STORM~\cite{huang2026storm} & 0.1756 & 0.9547 & 0.3229 & 0.8495 & 0.4673 & 0.6873 & 0.7074 & 0.2912 & 0.8017 & 0.1191 \\
\addlinespace[1.5pt]
\midrule[0.6pt]
\rowcolor{resultours}
\rule{0pt}{11.5pt}\textcolor{resultpurple}{\bfseries WeatherCraft (ours)} & \ranktwo{0.1453} & \ranktwo{0.9700} & \rankone{0.2586} & \rankone{0.9098} & \rankone{0.3721} & \rankone{0.8057} & \rankone{0.5474} & \rankone{0.4843} & \rankone{0.5967} & \rankone{0.2897}\\[1pt]
\addlinespace[1pt]
\textcolor{resultpurple}{\itshape Rel.\ gain vs.\ best baseline} & \textcolor{resultpurple}{-0.62\%} & \textcolor{resultpurple}{-0.03\%} & \textcolor{resultpurple}{+1.22\%} & \textcolor{resultpurple}{+0.63\%} & \textcolor{resultpurple}{+5.08\%} & \textcolor{resultpurple}{+2.43\%} & \textcolor{resultpurple}{+16.99\%} & \textcolor{resultpurple}{+22.79\%} & \textcolor{resultpurple}{+17.64\%} & \textcolor{resultpurple}{+55.25\%}\\[1pt]
\bottomrule[0.95pt]
\end{tabularx}
\end{table*}
\endgroup

\begin{figure*}[t]
\vspace{-4pt}
\centering
\includegraphics[width=\textwidth]{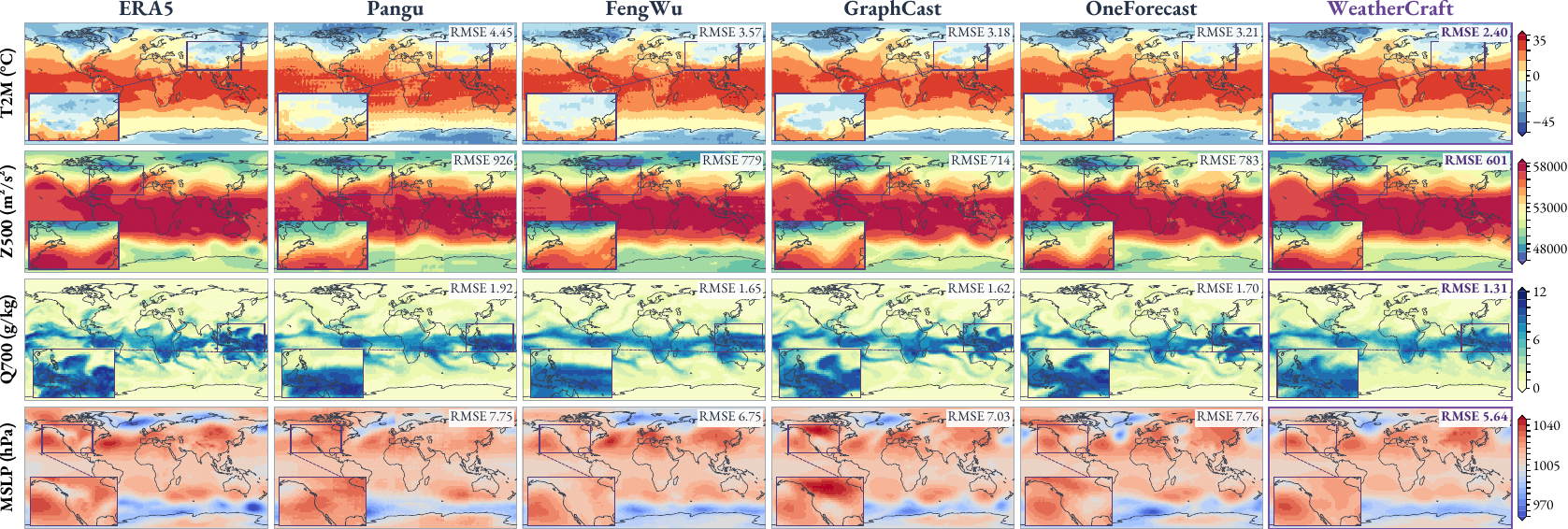}
\vspace{-10pt}
\caption{\textbf{Global $10$-day multivariable forecasts} (1 January 2020) across T2M, Z500, Q700, and MSLP. Deterministic baselines use single trajectories and \weathercraft{} uses an $8$-member mean, with regional insets (bottom left) and physical RMSE (top right).}
\label{fig:daily-fields}
\end{figure*}

% Side-by-side Table 2 (left) and Figure 4 (right) with matched top/bottom/caption baselines and EB Garamond styling.
\begingroup
\definecolor{resultink}{HTML}{292C42}
\definecolor{resultpurple}{HTML}{544178}
\definecolor{resultheader}{HTML}{E9E8F2}
\definecolor{resultstripe}{HTML}{F6F7FA}
\definecolor{resultours}{HTML}{EEE7F7}
\definecolor{rankfirst}{HTML}{E4D9F1}
\definecolor{ranksecond}{HTML}{F7E7D6}
\definecolor{rankthird}{HTML}{E0EDF4}
\newcommand{\rankone}[1]{\cellcolor{rankfirst}\textbf{#1}}
\newcommand{\ranktwo}[1]{\cellcolor{ranksecond}#1}
\newcommand{\rankthree}[1]{\cellcolor{rankthird}#1}

\begin{figure*}[t]
\vspace{-2pt}
\centering
\begin{minipage}[t]{0.482\textwidth}
\vspace{0pt}
\centering
\begingroup
\fontfamily{EBGaramond-LF}\fontsize{8.8}{10.8}\selectfont
\color{resultink}
\setlength{\tabcolsep}{1.5pt}
\renewcommand{\arraystretch}{1.50}
\setlength{\aboverulesep}{2.4pt}
\setlength{\belowrulesep}{2.4pt}
\begin{tabularx}{\linewidth}{@{}l*{5}{>{\centering\arraybackslash}X}r@{}}
\toprule[0.95pt]
\rowcolor{resultheader}
\rule{0pt}{10.5pt}\textbf{Method} & \textbf{1-d} & \textbf{3-d} & \textbf{5-d} & \textbf{10-d} & \textbf{15-d} & \multicolumn{1}{c@{}}{\textbf{Time$\downarrow$}} \\[-0.5pt]
\midrule[0.55pt]
\rowcolor{resultstripe} GenCast~\cite{price2025gencast} & \rankone{0.0510} & \rankone{0.0919} & \rankone{0.1347} & \rankone{0.2126} & \rankone{0.2400} & \,\,452.4\,s \\
Tyche~\cite{xu2026tyche} & \ranktwo{0.0515} & 0.1023 & 0.1527 & 0.2392 & 0.2729 & \rankone{\,\,7.2\,s} \\
\midrule[0.55pt]
\rowcolor{resultstripe} \weathercraft{} & 0.0602 & 0.1065 & 0.1526 & 0.2252 & 0.2476 & \ranktwo{\,\,25.2\,s} \\
\quad + $1$-day CRPS & \rankthree{0.0582} & \rankthree{0.1017} & \rankthree{0.1457} & \rankthree{0.2197} & \rankthree{0.2436} & \ranktwo{\,\,25.2\,s$^{*}$} \\
\rowcolor{resultours}
\rule{0pt}{11.0pt}\textcolor{resultpurple}{\bfseries \quad + $2$-day CRPS} & 0.0584 & \ranktwo{0.1008} & \ranktwo{0.1441} & \ranktwo{0.2181} & \ranktwo{0.2421} & \ranktwo{\,\,25.2\,s$^{*}$} \\
\addlinespace[1.0pt]
\multicolumn{6}{@{}l}{\textcolor{resultpurple}{\itshape Speedup vs.\ GenCast ($<1\%$ CRPS gap at $15$-d)}} & \textcolor{resultpurple}{\bfseries 17.9$\times$} \\
\bottomrule[0.95pt]
\end{tabularx}
\endgroup
\vspace{1.0pt}
\captionof{table}{\textbf{Ensemble fair-CRPS and $15$-day runtime.} Times are averaged over eight starts on one H100. $^{*}$Both tuned models use the pretrained model's measured runtime.}
\label{tab:probabilistic}
\end{minipage}%
\hfill
\begin{minipage}[t]{0.498\textwidth}
\vspace{0.0pt}
\centering
\includegraphics[width=\linewidth]{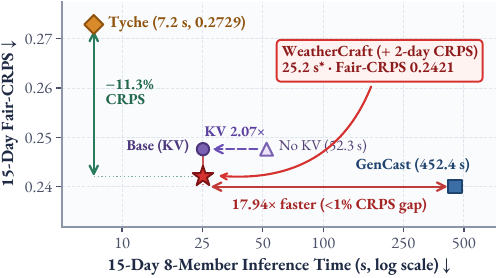}
\vspace{-16.73pt}
\caption{\textbf{Day-$15$ ensemble skill and inference cost.} Fair-CRPS versus runtime for eight members on one H100.}
\label{fig:inference-time}
\end{minipage}
\vspace{2pt}
\end{figure*}
\endgroup

\begin{figure*}[t]
\vspace{-2pt}
\centering
\includegraphics[width=\textwidth]{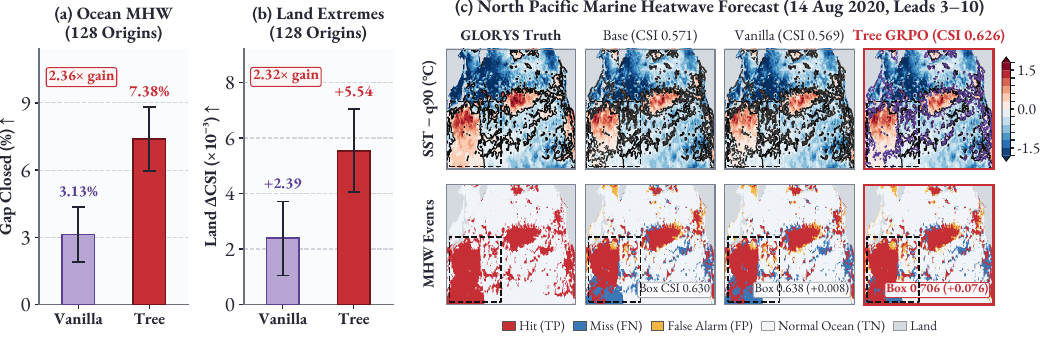}
\vspace{-14pt}
\caption{\textbf{Reward alignment for downstream extremes.} (a) Percentage of the Base-to-ERA5 forcing gap in SST exceedance CSI closed. (b) Persistent land-event CSI gain over Base ($10^{-3}$ units). Bars summarize $128$ test starts with $95\%$ CIs. (c) North Pacific marine heatwave case initialized on 14 August 2020, leads $3$--$10$.}
\label{fig:rl-gains}
\vspace{-4pt}
\end{figure*}

\begin{figure*}[t]
\vspace{-4pt}
\centering
\includegraphics[width=\textwidth]{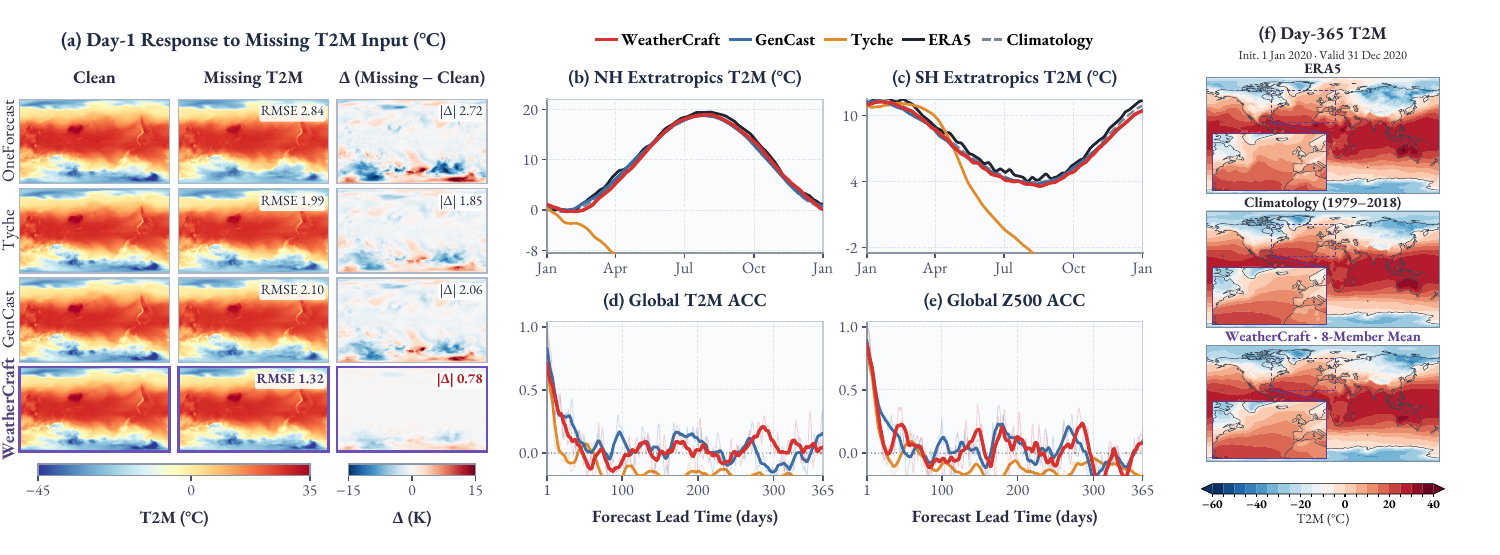}
\vspace{-14pt}
\caption{\textbf{Missing inputs and year-long rollouts.} (a) Day-$1$ T2M (1 January 2020) under climatology-imputed T2M ($\Delta=\widehat X_{\rm missing}-\widehat X_{\rm clean}$). (b--e) $365$-day rollouts (1 January 2019). (f) Separate day-$365$ T2M rollout (1 January 2020; insets: $80^\circ\text{W}$--$20^\circ\text{E}$, $20^\circ$--$70^\circ\text{N}$).}
\label{fig:missing-fields}
\label{fig:annual}
\vspace{1.5pt}
\end{figure*}

\subsection{Experimental Setup}
\label{sec:exp-setup}

\paragraph{Data and baselines.}
We evaluate on ERA5~\cite{hersbach2020era5} global $1.5^\circ$ daily-mean fields ($120\times 240$ grid, $69$ channels), split into 1979--2018 for training, 2019 for validation, and 2020--2022 for testing ($1{,}081$ initializations, $15$-day rollouts). All baselines are retrained on the same data, including deterministic models ConvLSTM~\cite{shi2015convlstm}, SimVP~\cite{gao2022simvp}, SFNO~\cite{bonev2023sfno}, Pangu-Weather~\cite{bi2023pangu}, OneForecast~\cite{gao2025oneforecast}, FuXi~\cite{chen2023fuxi}, FengWu~\cite{chen2025fengwu}, GraphCast~\cite{lam2023graphcast}, and STORM~\cite{huang2026storm}, as well as probabilistic generators GenCast~\cite{price2025gencast} and Tyche~\cite{xu2026tyche}. Probabilistic models generate eight ensemble members (evaluating ensemble means for point forecasts), whereas deterministic baselines output a single forecast trajectory.

\paragraph{Implementation and metrics.}
\weathercraft{} uses a $16$-layer block-causal Transformer ($1{,}536$ width, $24$ heads, ${\sim}460\text{M}$ parameters), trained on $8\times$ NVIDIA H100 GPUs in \texttt{bfloat16} ($50$ pretraining epochs at $2\times 10^{-4}$ peak learning rate; post-training at $10^{-6}$, with \texttt{float32} lookup and decoding for CRPS tuning). Runtime benchmarks use one H100 GPU with each model's native framework and numerical precision. We report latitude-weighted normalized RMSE (nRMSE), anomaly correlation coefficient (ACC), and unbiased fair-CRPS~\cite{ferro2014fairscores} across all $69$ standardized channels (full details in Appendices~\ref{app:architecture}--\ref{app:eval-ablation}).

\subsection{Daily Weather Forecasting}
\label{sec:exp-daily}

\paragraph{Quantitative comparison.}
Table~\ref{tab:daily} compares \weathercraft{} against nine deterministic baselines over $1$--$15$-day leads. At day $1$, \weathercraft{} ($\mathrm{nRMSE}=0.1453$, $\mathrm{ACC}=0.9700$) closely matches OneForecast ($0.1444$, $0.9703$), showing competitive short-range point skill (tokenizer reconstruction is evaluated separately in Appendix~\ref{app:b-reconstruction}). From day $3$ onward, \weathercraft{} achieves the best nRMSE and ACC at every reported lead, reducing nRMSE over OneForecast by $5.08\%$ at day $5$ ($0.3721$ vs.\ $0.3920$) and $16.99\%$ at day $10$ ($0.5474$ vs.\ $0.6594$). At day $15$, \weathercraft{} reaches $0.5967$ nRMSE ($17.64\%$ below SFNO's $0.7245$) and $0.2897$ ACC ($0.1031$ above OneForecast's $0.1866$), surpassing the strongest deterministic baseline on each respective evaluation metric (Appendix Tables~\ref{tab:d-variable-rmse} and~\ref{tab:d-variable-acc}).

\paragraph{Qualitative comparison.}
Figure~\ref{fig:daily-fields} compares $10$-day global forecasts initialized on 1 January 2020 alongside zoomed-in regional insets. For this initialization, deterministic regression forecasts exhibit smoother spatial structures or larger phase displacements across several synoptic systems. By contrast, the \weathercraft{} ensemble mean more closely preserves the ERA5 Siberian temperature fronts (T2M), North Atlantic trough--ridge wave trains (Z500), tropical moisture plumes (Q700), and extratropical cyclone dipoles (MSLP), achieving the lowest physical-unit RMSE among the displayed methods on all four variables.

\subsection{Probabilistic Forecasting and Efficiency}
\label{sec:exp-probabilistic}

\paragraph{Probabilistic skill.}
Pretrained \weathercraft{} achieves lower fair-CRPS than Tyche at days $5$, $10$, and $15$ (Table~\ref{tab:probabilistic}). Sequential one- and two-day fair-CRPS tuning improves every reported lead over pretraining (Sec.~\ref{app:c-crps}). At day $15$, the tuned model reaches $0.2421$, within $1\%$ of GenCast ($0.2400$) and $11.3\%$ below Tyche ($0.2729$). Its ensemble-mean nRMSE also improves to $0.5906$ (Appendix Table~\ref{tab:c-crps}).

\paragraph{Accuracy--efficiency trade-off.}
On one H100, block KV caching cuts the pretrained model's eight-member, $15$-day rollout from $52.26$ to $25.22$\,s ($2.07\times$ speedup). The cached model runs $17.9\times$ faster than GenCast ($452.38$\,s at $39$ steps per day). Tyche completes the same rollout in $7.22$\,s (Fig.~\ref{fig:inference-time}).

\subsection{Reinforcement Learning for Downstream Tasks}
\label{sec:exp-rl}

\paragraph{Quantitative gains.}
We reward forecasts using the critical success index ($\mathrm{CSI}=\mathrm{TP}/(\mathrm{TP}+\mathrm{FP}+\mathrm{FN})$). The ocean task scores daily SST exceedances above the calendar-day 90th percentile at leads $3$--$10$ under a frozen TritonCast model~\cite{wu2025tritoncast}. The land task scores temperature extremes lasting at least three days (Sec.~\ref{app:c-downstream}). Across $128$ test initializations, Tree GRPO raises mean ocean CSI from $0.55605$ to $0.55979$, closing $7.38\%$ of the gap to ERA5 forcing versus $3.13\%$ for Vanilla GRPO. Mean land CSI improves by $0.00554$ and $0.00239$, respectively. Tree GRPO thus delivers over $2.3\times$ the gain of Vanilla GRPO on both tasks (Fig.~\ref{fig:rl-gains}(a,b)).

\paragraph{Coupled marine heatwave analysis.}
In the coupled North Pacific marine heatwave initialized on 14 August 2020 (Fig.~\ref{fig:rl-gains}(c)), the base and Vanilla GRPO policies suffer from extensive misses (blue) in the western subtropical core (dashed box), whereas Tree GRPO closely tracks the observed GLORYS boundary and converts these misses into hits (red), boosting regional CSI from $0.571$ to $0.626$ and focal-box CSI from $0.630$ to $0.706$ ($+0.076$).

\subsection{Generalization, Robustness, and Ablations}
\label{sec:exp-generalization}

\paragraph{Missing input variables.}
Replacing one initial variable at a time (T2M, MSLP, or U10) with calendar-day climatology over $32$ test cases and aggregating squared errors across all daily leads $1$--$15$ before taking the square root, \weathercraft{} maintains nearly unchanged omitted-channel RMSE between clean and missing inputs ($2.0560$ vs.\ $2.0547$\,K for T2M; $5.2088$ vs.\ $5.1801$\,hPa for MSLP; $2.8408$ vs.\ $2.8506$\,m\,s$^{-1}$ for U10), with the lowest day-$1$ T2M perturbation ($|\Delta|=0.78$\,K; Fig.~\ref{fig:missing-fields}(a)). A single pass through the frozen tokenizer reduces missing-channel reconstruction RMSE relative to climatology-imputed inputs by $44.6\%$--$76.6\%$ across variables (Appendix Tables~\ref{tab:d-missing} and~\ref{tab:d-tokenizer-recovery}).

\paragraph{Annual rollout.}
In $365$-day rollouts from 1 January 2019 (Fig.~\ref{fig:annual}(b--e)), daily spatial ACC decays toward zero after two weeks under chaotic divergence (Fig.~\ref{fig:annual}(d,e)), while \weathercraft{} and GenCast retain the broad extratropical seasonal temperature cycle across both hemispheres whereas Tyche develops a pronounced cold bias (Fig.~\ref{fig:annual}(b,c)). In a separate $365$-day rollout from 1 January 2020 (Fig.~\ref{fig:annual}(f)), the day-$365$ \weathercraft{} mean stays physically stable and closely matches climatology ($1.45^\circ\text{C}$ RMSE) across land--sea and North Atlantic gradients.

\paragraph{Design ablations.}
Controlled ablations in Appendix~\ref{app:eval-ablation} validate three core architectural choices: (1) \textit{residual FSQ vs.\ grouped VQ}: under matched capacity, residual FSQ outperforms VQ in both nRMSE (day-$5$ $0.3721$ vs.\ $0.3791$) and fair-CRPS across $1$--$10$-day leads (Table~\ref{tab:tokenizer_forecast}); (2) \textit{residual vs.\ independent multiscale quantization}: residual quantization cuts test reconstruction nRMSE by $19.9\%$ from $0.0828$ to $0.0664$ across all $69$ channels (Table~\ref{tab:tokenizer_reconstruction}); and (3) \textit{within-scale parallel decoding}: parallel decoding achieves lower error (day-$5$ nRMSE $0.3707$ vs.\ $0.3791$) and fair-CRPS than $5\times 10$ spatial-block autoregression (Table~\ref{tab:spatial_block}).

\vspace{6.9pt}
\section{Conclusion}
\label{sec:conclusion}

We presented \weathercraft{}, a unified framework combining discrete multiscale weather tokenization, block-causal ensemble generation, and reward-aligned post-training without differentiating through downstream evaluators. Combining multi-group residual FSQ with scale-parallel latent rollouts and block KV caching delivers strong medium-range point and probabilistic skill on global $69$-channel ERA5 benchmarks at $17.9\times$ the speed of GenCast over eight-member, $15$-day rollouts. Sequential fair-CRPS tuning sharpens ensemble calibration across leads, while rollout-tree GRPO uses exact token log-probabilities to optimize non-differentiable persistent land-extreme and coupled-ocean heatwave CSI rewards alongside robust missing-input recovery and stable year-long rollouts.

More broadly, discrete autoregressive policies bridge general-purpose atmospheric pretraining and downstream Earth system tasks whose evaluators or coupled simulators are costly or non-differentiable. Future work will scale multiscale tokenization to finer spatial--temporal resolutions, assimilate heterogeneous multimodal observations, and extend reward alignment to operational multi-hazard warnings and decision-making under physical constraints.

\FloatBarrier
{\footnotesize
\bibliographystyle{ieeenat_fullname}
\bibliography{main_ref}

@article{bi2023pangu,
  title       = {{Accurate medium-range global weather forecasting with 3D neural networks}},
  author      = {Kaifeng Bi and Lingxi Xie and Hengheng Zhang and Xin Chen and Xiaotao Gu and Qi Tian},
  year        = {2023},
  journal     = {Nature},
  doi         = {10.1038/s41586-023-06185-3},
  volume      = {619},
  pages       = {533--538}
}

@article{lam2023graphcast,
  title       = {{Learning skillful medium-range global weather forecasting}},
  author      = {Remi Lam and Alvaro Sanchez-Gonzalez and Matthew Willson and Peter Wirnsberger and Meire Fortunato and Ferran Alet and Suman Ravuri and Timo Ewalds and Zach Eaton-Rosen and Weihua Hu and Alexander Merose and Stephan Hoyer and George Holland and Oriol Vinyals and Jacklynn Stott and Alexander Pritzel and Shakir Mohamed and Peter Battaglia},
  year        = {2023},
  journal     = {Science},
  doi         = {10.1126/science.adi2336},
  volume      = {382},
  number      = {6677},
  pages       = {1416--1421}
}

@article{chen2023fuxi,
  title       = {{FuXi: a cascade machine learning forecasting system for 15-day global weather forecast}},
  author      = {Lei Chen and Xiaohui Zhong and Feng Zhang and Yuan Cheng and Yinghui Xu and Yuan Qi and Hao Li},
  year        = {2023},
  journal     = {npj Climate and Atmospheric Science},
  doi         = {10.1038/s41612-023-00512-1},
  volume      = {6},
  pages       = {190}
}

@article{chen2025fengwu,
  title       = {{The operational medium-range deterministic weather forecasting can be extended beyond a 10-day lead time}},
  author      = {Kang Chen and Tao Han and Junchao Gong and Lei Bai and Fenghua Ling and Jing-Jia Luo and Xi Chen and Leiming Ma and Tianning Zhang and Rui Su and Yuanzheng Ci and Bin Li and Xiaokang Yang and Wanli Ouyang},
  year        = {2025},
  journal     = {Communications Earth \& Environment},
  doi         = {10.1038/s43247-025-02502-y},
  volume      = {6},
  pages       = {518}
}

@misc{pathak2022fourcastnet,
  title       = {{FourCastNet: A Global Data-driven High-resolution Weather Model using Adaptive Fourier Neural Operators}},
  author      = {Jaideep Pathak and Shashank Subramanian and Peter Harrington and Sanjeev Raja and Ashesh Chattopadhyay and Morteza Mardani and Thorsten Kurth and David Hall and Zongyi Li and Kamyar Azizzadenesheli and Pedram Hassanzadeh and Karthik Kashinath and Anima Anandkumar},
  year        = {2022},
  url         = {https://arxiv.org/abs/2202.11214},
  note        = {arXiv:2202.11214}
}

@inproceedings{gao2025oneforecast,
  title       = {{OneForecast: A Universal Framework for Global and Regional Weather Forecasting}},
  author      = {Yuan Gao and Hao Wu and Ruiqi Shu and Huanshuo Dong and Fan Xu and Rui Ray Chen and Yibo Yan and Qingsong Wen and Xuming Hu and Kun Wang and Jiahao Wu and Li Qing and Hui Xiong and Xiaomeng Huang},
  year        = {2025},
  booktitle   = {International Conference on Machine Learning},
  url         = {https://proceedings.mlr.press/v267/gao25r.html},
  volume      = {267},
  pages       = {18658--18697}
}

@inproceedings{huang2026storm,
 author = {Huang, Qihe and Zhou, Zhengyang and Li, Yangze and Ma, Jiaming and Yang, Kuo and Wang, Binwu and Wang, Xu and Wang, Yang},
 booktitle = {International Conference on Learning Representations},
 editor = {C. Vondrick and B. Hariharan and C. Raffel and L. Pinto and D. Yang and A. Faust},
 pages = {95838--95870},
 title = {STORM: Synergistic Cross-Scale Spatio-Temporal Modeling for Weather Forecasting},
 url = {https://proceedings.iclr.cc/paper_files/paper/2026/file/9ad996b5c45130de2bc00b60d8607904-Paper-Conference.pdf},
 volume = {2026},
 year = {2026}
}

@inproceedings{shi2015convlstm,
  title       = {{Convolutional LSTM Network: A Machine Learning Approach for Precipitation Nowcasting}},
  author      = {Xingjian Shi and Zhourong Chen and Hao Wang and Dit-Yan Yeung and Wai-Kin Wong and Wang-Chun Woo},
  year        = {2015},
  booktitle   = {Advances in Neural Information Processing Systems},
  url         = {https://proceedings.neurips.cc/paper/2015/hash/07563a3fe3bbe7e3ba84431ad9d055af-Abstract.html},
  volume      = {28}
}

@inproceedings{gao2022simvp,
  title       = {{SimVP: Simpler Yet Better Video Prediction}},
  author      = {Zhangyang Gao and Cheng Tan and Lirong Wu and Stan Z. Li},
  year        = {2022},
  booktitle   = {IEEE/CVF Conference on Computer Vision and Pattern Recognition},
  doi         = {10.1109/CVPR52688.2022.00317},
  pages       = {3170--3180}
}

@inproceedings{bonev2023sfno,
  title       = {{Spherical Fourier Neural Operators: Learning Stable Dynamics on the Sphere}},
  author      = {Boris Bonev and Thorsten Kurth and Christian Hundt and Jaideep Pathak and Maximilian Baust and Karthik Kashinath and Anima Anandkumar},
  year        = {2023},
  booktitle   = {International Conference on Machine Learning},
  url         = {https://proceedings.mlr.press/v202/bonev23a.html},
  volume      = {202},
  pages       = {2806--2823}
}

@misc{wu2025tritoncast,
  title       = {{Advanced Long-term Earth System Forecasting}},
  author      = {Hao Wu and Yuan Gao and Ruijian Gou and Xian Wu and Chuhan Wu and Huahui Yi and Johannes Brandstetter and Fan Xu and Kun Wang and Penghao Zhao and Hao Jia and Qi Song and Xinliang Liu and Juncai He and Shuhao Cao and Huanshuo Dong and Yanfei Xiang and Fan Zhang and Haixin Wang and Xingjian Shi and Qiufeng Wang and Shuaipeng Li and Ruobing Xie and Feng Tao and Yuxu Lu and Yu Guo and Yuntian Chen and Yuxuan Liang and Qingsong Wen and Wanli Ouyang and Deliang Chen and Niklas Boers and Xiaomeng Huang},
  year        = {2025},
  url         = {https://arxiv.org/abs/2505.19432},
  note        = {arXiv:2505.19432, revised 2026}
}

@article{price2025gencast,
  title       = {{Probabilistic weather forecasting with machine learning}},
  author      = {Ilan Price and Alvaro Sanchez-Gonzalez and Ferran Alet and Tom R. Andersson and Andrew El-Kadi and Dominic Masters and Timo Ewalds and Jacklynn Stott and Shakir Mohamed and Peter Battaglia and Remi Lam and Matthew Willson},
  year        = {2025},
  journal     = {Nature},
  doi         = {10.1038/s41586-024-08252-9},
  volume      = {637},
  pages       = {84--90}
}

@article{couairon2026archesweathergen,
  title       = {{ArchesWeatherGen: Skillful and compute-efficient probabilistic weather forecasting with machine learning}},
  author      = {Guillaume Couairon and Renu Singh and Anastase Charantonis and Christian Lessig and Claire Monteleoni},
  year        = {2026},
  journal     = {Science Advances},
  doi         = {10.1126/sciadv.adx2372},
  volume      = {12},
  number      = {17},
  pages       = {eadx2372}
}

@misc{xu2026tyche,
  title       = {{Tyche: One Step Flow for Efficient Probabilistic Weather Forecasting}},
  author      = {Fan Xu and Yuan Gao and Kun Wang and Rui Su and Fenghua Ling and Hao Wu and Wanli Ouyang},
  year        = {2026},
  url         = {https://arxiv.org/abs/2605.06916},
  note        = {arXiv:2605.06916}
}

@inproceedings{zhdanov2026mosaic,
  title       = {{(Sparse) Attention to the Details: Preserving Spectral Fidelity in ML-based Weather Forecasting Models}},
  author      = {Maksim Zhdanov and Ana Lucic and Max Welling and Jan-Willem van de Meent},
  year        = {2026},
  booktitle   = {International Conference on Machine Learning},
  url         = {https://proceedings.mlr.press/v306/zhdanov26a.html},
  volume      = {306},
  pages       = {163086--163115}
}

@article{alet2026cyclones,
  title       = {{Operational tropical cyclone forecasting with AI}},
  author      = {Ferran Alet and Tom R. Andersson and Ilan Price and Stratis Markou and Andrew El-Kadi and Dominic Masters and Amy Li and Samier Merchant and Natalie Williams and Gregory Thornton and Ken MacKay and Olivia Graham and Akib Uddin and Ben Gaiarin and Devaja Shah and Elinor Kruse and Wallace Hogsett and David Zelinsky and John Cangialosi and Jonathan Martinez and James Franklin and Mark DeMaria and Kate Musgrave and Caroline L. Bain and Helen Titley and Jacklynn Stott and Remi Lam and Aaron Bell and Paul Komarek and Matthew Willson and Alvaro Sanchez-Gonzalez and Peter Battaglia},
  year        = {2026},
  journal     = {Nature},
  doi         = {10.1038/s41586-026-10953-2},
  volume      = {657},
  number      = {8132},
  pages       = {680--688}
}

@article{lang2026aifscrps,
  title       = {{AIFS-CRPS: ensemble forecasting using a model trained with a loss function based on the continuous ranked probability score}},
  author      = {Simon Lang and Mihai Alexe and Mariana C. A. Clare and Christopher Roberts and Rilwan Adewoyin and Zied Ben Bouallegue and Matthew Chantry and Jesper Dramsch and Peter D. Dueben and Sara Hahner and Pedro Maciel and Ana Prieto-Nemesio and Cathal OBrien and Florian Pinault and Jan Polster and Baudouin Raoult and Steffen Tietsche and Martin Leutbecher},
  year        = {2026},
  journal     = {npj Artificial Intelligence},
  doi         = {10.1038/s44387-026-00073-7},
  volume      = {2},
  pages       = {18}
}

@article{ferro2014fairscores,
  title       = {{Fair scores for ensemble forecasts}},
  author      = {C. A. T. Ferro},
  year        = {2014},
  journal     = {Quarterly Journal of the Royal Meteorological Society},
  doi         = {10.1002/qj.2270},
  volume      = {140},
  number      = {683},
  pages       = {1917--1923}
}

@article{stock2026swift,
  author = {Stock, Jason and Arcomano, Troy and Kotamarthi, Rao},
  title = {{Swift}: An Autoregressive Consistency Model for Efficient Weather Forecasting},
  journal = {Machine Learning: Earth},
  year = {2026},
  volume = {2},
  pages = {025004},
  doi = {10.1088/3049-4753/ae802a}
}

@article{rasp2024weatherbench2,
  title       = {{WeatherBench 2: A Benchmark for the Next Generation of Data-Driven Global Weather Models}},
  author      = {Stephan Rasp and Stephan Hoyer and Alexander Merose and Ian Langmore and Peter Battaglia and Tyler Russell and Alvaro Sanchez-Gonzalez and Vivian Yang and Rob Carver and Shreya Agrawal and Matthew Chantry and Zied Ben Bouallegue and Peter Dueben and Carla Bromberg and Jared Sisk and Luke Barrington and Aaron Bell and Fei Sha},
  year        = {2024},
  journal     = {Journal of Advances in Modeling Earth Systems},
  doi         = {10.1029/2023MS004019},
  volume      = {16},
  number      = {6},
  pages       = {e2023MS004019}
}

@inproceedings{oord2017vqvae,
  title       = {{Neural Discrete Representation Learning}},
  author      = {A{\"a}ron van den Oord and Oriol Vinyals and Koray Kavukcuoglu},
  year        = {2017},
  booktitle   = {Advances in Neural Information Processing Systems},
  url         = {https://proceedings.neurips.cc/paper_files/paper/2017/hash/7a98af17e63a0ac09ce2e96d03992fbc-Abstract.html},
  volume      = {30}
}

@article{bengio2013ste,
  author = {Yoshua Bengio and Nicholas L{\'{e}}onard and Aaron C. Courville},
  title = {Estimating or Propagating Gradients Through Stochastic Neurons for Conditional Computation},
  journal = {CoRR},
  volume = {abs/1308.3432},
  year = {2013},
  url = {https://arxiv.org/abs/1308.3432},
  eprinttype = {arXiv},
  eprint = {1308.3432},
  bibsource = {dblp computer science bibliography, https://dblp.org}
}

@inproceedings{esser2021vqgan,
  title       = {{Taming Transformers for High-Resolution Image Synthesis}},
  author      = {Patrick Esser and Robin Rombach and Bj{\"o}rn Ommer},
  year        = {2021},
  booktitle   = {IEEE/CVF Conference on Computer Vision and Pattern Recognition},
  url         = {https://openaccess.thecvf.com/content/CVPR2021/html/Esser_Taming_Transformers_for_High-Resolution_Image_Synthesis_CVPR_2021_paper.html},
  pages       = {12873--12883}
}

@inproceedings{mentzer2024fsq,
  title       = {{Finite Scalar Quantization: VQ-VAE Made Simple}},
  author      = {Fabian Mentzer and David Minnen and Eirikur Agustsson and Michael Tschannen},
  year        = {2024},
  booktitle   = {International Conference on Learning Representations},
  url         = {https://proceedings.iclr.cc/paper_files/paper/2024/hash/e2dd53601de57c773343a7cdf09fae1c-Abstract-Conference.html}
}

@inproceedings{yu2024lfq,
  title       = {{Language Model Beats Diffusion -- Tokenizer Is Key to Visual Generation}},
  author      = {Lijun Yu and Jose Lezama and Nitesh Bharadwaj Gundavarapu and Luca Versari and Kihyuk Sohn and David Minnen and Yong Cheng and Agrim Gupta and Xiuye Gu and Alexander G. Hauptmann and Boqing Gong and Ming-Hsuan Yang and Irfan Essa and David Ross and Lu Jiang},
  year        = {2024},
  booktitle   = {International Conference on Learning Representations},
  url         = {https://proceedings.iclr.cc/paper_files/paper/2024/hash/036912a83bdbb1fd792baf6532f102d8-Abstract-Conference.html}
}

@misc{lingsch2026phaedra,
  title       = {{Phaedra: Learning High-Fidelity Discrete Tokenization for the Physical Science}},
  author      = {Levi Lingsch and Georgios Kissas and Johannes Jakubik and Siddhartha Mishra},
  year        = {2026},
  eprint      = {2602.03915},
  archivePrefix = {arXiv},
  primaryClass = {cs.CV},
  url         = {https://arxiv.org/abs/2602.03915},
  note        = {arXiv preprint arXiv:2602.03915}
}

@misc{mirowski2024neuralcompressionatmosphericstates,
      title={Neural Compression of Atmospheric States}, 
      author={Piotr Mirowski and David Warde-Farley and Mihaela Rosca and Matthew Koichi Grimes and Yana Hasson and Hyunjik Kim and Mélanie Rey and Simon Osindero and Suman Ravuri and Shakir Mohamed},
      year={2024},
      eprint={2407.11666},
      archivePrefix={arXiv},
      primaryClass={cs.LG},
      url={https://arxiv.org/abs/2407.11666}, 
      note={arXiv:2407.11666},
}

@misc{zhao2025transformingweatherdatapixel,
      title={Transforming Weather Data from Pixel to Latent Space}, 
      author={Sijie Zhao and Feng Liu and Xueliang Zhang and Hao Chen and Tao Han and Junchao Gong and Ran Tao and Pengfeng Xiao and Lei Bai and Wanli Ouyang},
      year={2025},
      eprint={2503.06623},
      archivePrefix={arXiv},
      primaryClass={cs.CV},
      url={https://arxiv.org/abs/2503.06623}, 
      note={arXiv:2503.06623},
}

@inproceedings{lee2022rq,
  title       = {{Autoregressive Image Generation Using Residual Quantization}},
  author      = {Doyup Lee and Chiheon Kim and Saehoon Kim and Minsu Cho and Wook-Shin Han},
  year        = {2022},
  booktitle   = {IEEE/CVF Conference on Computer Vision and Pattern Recognition},
  doi         = {10.1109/CVPR52688.2022.01123},
  pages       = {11523--11532}
}

@inproceedings{yu2026climatear,
  title       = {{ClimateAR: Multi-Scale Autoregressive Generative Modeling for Climate Forecasting}},
  author      = {Yue Yu and Weiqi Chen and Binqing Wu and Dongliang Cui and Wanyi Jiang and Zongjiang Shang and Bo Wu and Liang Sun and Ling Chen},
  year        = {2026},
  booktitle   = {International Conference on Machine Learning},
  url         = {https://proceedings.mlr.press/v306/yu26f.html},
  volume      = {306},
  pages       = {150113--150140}
}

@article{franch2025gptcast,
  title       = {{GPTCast: a weather language model for precipitation nowcasting}},
  author      = {Gabriele Franch and Elena Tomasi and Rishabh Wanjari and Virginia Poli and Chiara Cardinali and Pier Paolo Alberoni and Marco Cristoforetti},
  year        = {2025},
  journal     = {Geoscientific Model Development},
  doi         = {10.5194/gmd-18-5351-2025},
  volume      = {18},
  pages       = {5351--5371}
}

@misc{wang2025beamvq,
  title       = {{BeamVQ: Beam Search with Vector Quantization to Mitigate Data Scarcity in Physical Spatiotemporal Forecasting}},
  author      = {Weiyan Wang and Xingjian Shi and Ruiqi Shu and Yuan Gao and Rui Ray Chen and Kun Wang and Fan Xu and Jinbao Xue and Shuaipeng Li and Yangyu Tao and Di Wang and Hao Wu and Xiaomeng Huang},
  year        = {2025},
  url         = {https://arxiv.org/abs/2502.18925},
  note        = {arXiv:2502.18925}
}

@inproceedings{ho2020ddpm,
  title       = {{Denoising Diffusion Probabilistic Models}},
  author      = {Jonathan Ho and Ajay N. Jain and Pieter Abbeel},
  year        = {2020},
  booktitle   = {Advances in Neural Information Processing Systems},
  url         = {https://proceedings.neurips.cc/paper/2020/hash/4c5bcfec8584af0d967f1ab10179ca4b-Abstract.html},
  volume      = {33}
}

@techreport{radford2019gpt2,
  title       = {{Language Models are Unsupervised Multitask Learners}},
  author      = {Alec Radford and Jeffrey Wu and Rewon Child and David Luan and Dario Amodei and Ilya Sutskever},
  year        = {2019},
  institution = {OpenAI},
  url         = {https://cdn.openai.com/better-language-models/language_models_are_unsupervised_multitask_learners.pdf}
}

@inproceedings{tian2024var,
  title       = {{Visual Autoregressive Modeling: Scalable Image Generation via Next-Scale Prediction}},
  author      = {Keyu Tian and Yi Jiang and Zehuan Yuan and Bingyue Peng and Liwei Wang},
  year        = {2024},
  booktitle   = {Advances in Neural Information Processing Systems},
  doi         = {10.52202/079017-2694},
  volume      = {37}
}

@misc{gallici2025vargrpo,
  title       = {{Fine-Tuning Next-Scale Visual Autoregressive Models with Group Relative Policy Optimization}},
  author      = {Matteo Gallici and Haitz {S{\'a}ez de Oc{\'a}riz Borde}},
  year        = {2025},
  eprint      = {2505.23331},
  archivePrefix = {arXiv},
  primaryClass = {cs.CV},
  url         = {https://arxiv.org/abs/2505.23331},
  note        = {arXiv preprint arXiv:2505.23331}
}

@inproceedings{li2026logitrefiner,
  title       = {{Logit Refiner: Improving Visual Autoregressive Models via Intra-Scale Dependency Modeling}},
  author      = {Meimingwei Li and Stefan Andreas Baumann and Felix Krause and Bj{\"o}rn Ommer},
  year        = {2026},
  booktitle   = {Computer Vision -- ECCV 2026},
  pages       = {447--464},
  publisher   = {Springer},
  url         = {https://link.springer.com/book/10.1007/978-3-032-37029-7}
}

@misc{shao2024grpo,
  title       = {{DeepSeekMath: Pushing the Limits of Mathematical Reasoning in Open Language Models}},
  author      = {Zhihong Shao and Peiyi Wang and Qihao Zhu and Runxin Xu and Junxiao Song and Xiao Bi and Haowei Zhang and Mingchuan Zhang and Y. K. Li and Y. Wu and Daya Guo},
  year        = {2024},
  url         = {https://arxiv.org/abs/2402.03300},
  note        = {arXiv:2402.03300}
}

@misc{ji2025treegrpo,
  title         = {{Tree Search for LLM Agent Reinforcement Learning}},
  author        = {Yuxiang Ji and Ziyu Ma and Yong Wang and Guanhua Chen and Xiangxiang Chu and Liaoni Wu},
  year          = {2025},
  eprint        = {2509.21240},
  archivePrefix = {arXiv},
  primaryClass  = {cs.LG},
  url           = {https://arxiv.org/abs/2509.21240},
  note          = {arXiv preprint arXiv:2509.21240}
}

@article{hersbach2020era5,
  title       = {{The ERA5 global reanalysis}},
  author      = {Hans Hersbach and Bill Bell and Paul Berrisford and Shoji Hirahara and Andras Horanyi and Joaquin Munoz-Sabater and Julien Nicolas and Carole Peubey and Raluca Radu and Dinand Schepers and Adrian Simmons and Cornel Soci and Saleh Abdalla and Xabier Abellan and Gianpaolo Balsamo and Peter Bechtold and Gionata Biavati and Jean Bidlot and Massimo Bonavita and Giovanni De Chiara and Per Dahlgren and Dick Dee and Michail Diamantakis and Rossana Dragani and Jacqueline Flemming and Richard Forbes and Manuel Fuentes and Alan Geer and Leo Haimberger and Sean Healy and Robin J. Hogan and Elizaveta Holm and Malgorzata Janiskova and Sarah Keeley and Patrick Laloyaux and Philippe Lopez and Cristina Lupu and Gabor Radnoti and Patricia de Rosnay and Iryna Rozum and Freja Vamborg and Sebastien Villaume and Jean-Noel Thepaut},
  year        = {2020},
  journal     = {Quarterly Journal of the Royal Meteorological Society},
  doi         = {10.1002/qj.3803},
  volume      = {146},
  number      = {730},
  pages       = {1999--2049}
}

@article{shu2025advancedmhw,
  title       = {{Advanced forecasts of global extreme marine heatwaves through a physics-guided data-driven approach}},
  author      = {Ruiqi Shu and Hao Wu and Yuan Gao and Fanghua Xu and Ruijian Gou and Wei Xiong and Xiaomeng Huang},
  year        = {2025},
  journal     = {Environmental Research Letters},
  volume      = {20},
  number      = {4},
  pages       = {044030},
  doi         = {10.1088/1748-9326/adbddd}
}

@article{lorenz1963chaos,
  author = {Edward N. Lorenz},
  title = {{Deterministic Nonperiodic Flow}},
  journal = {Journal of the Atmospheric Sciences},
  year = {1963},
  volume = {20},
  number = {2},
  pages = {130--141},
  doi = {10.1175/1520-0469(1963)020<0130:DNF>2.0.CO;2}
}

@article{leutbecher2008ensemble,
  author = {Martin Leutbecher and Tim N. Palmer},
  title = {{Ensemble Forecasting}},
  journal = {Journal of Computational Physics},
  year = {2008},
  volume = {227},
  number = {7},
  pages = {3515--3539},
  doi = {10.1016/j.jcp.2007.02.014}
}

@inproceedings{lipman2023flowmatching,
  author = {Yaron Lipman and Ricky T. Q. Chen and Heli Ben-Hamu and Maximilian Nickel and Matt Le},
  title = {{Flow Matching for Generative Modeling}},
  booktitle = {International Conference on Learning Representations},
  year = {2023},
  url = {https://arxiv.org/abs/2210.02747}
}

@inproceedings{ouyang2022instructgpt,
  author = {Long Ouyang and Jeff Wu and Xu Jiang and Diogo Almeida and Carroll L. Wainwright and Pamela Mishkin and Chong Zhang and Sandhini Agarwal and Katarina Slama and Alex Ray and John Schulman and Jacob Hilton and Fraser Kelton and Luke Miller and Maddie Simens and Amanda Askell and Peter Welinder and Paul Christiano and Jan Leike and Ryan Lowe},
  title = {{Training Language Models to Follow Instructions with Human Feedback}},
  booktitle = {Advances in Neural Information Processing Systems},
  year = {2022},
  volume = {35},
  pages = {27730--27744},
  url = {https://arxiv.org/abs/2203.02155}
}

@article{williams1992reinforce,
  author = {Ronald J. Williams},
  title = {{Simple Statistical Gradient-Following Algorithms for Connectionist Reinforcement Learning}},
  journal = {Machine Learning},
  year = {1992},
  volume = {8},
  pages = {229--256},
  doi = {10.1007/BF00992696}
}

@article{zhong2025fuxiens,
  title = {{FuXi-ENS: A machine learning model for efficient and accurate ensemble weather prediction}},
  author = {Xiaohui Zhong and Lei Chen and Hao Li and Roberto Buizza and Jun Liu and Jie Feng and Zijian Zhu and Xu Fan and Kan Dai and Jing-jia Luo and Jie Wu and Bo Lu},
  journal = {Science Advances},
  year = {2025},
  volume = {11},
  number = {44},
  pages = {eadu2854},
  doi = {10.1126/sciadv.adu2854}
}

@misc{alet2025fgn,
  title = {{Skillful joint probabilistic weather forecasting from marginals}},
  author = {Ferran Alet and Ilan Price and Andrew El-Kadi and Dominic Masters and Stratis Markou and Tom R. Andersson and Jacklynn Stott and Remi Lam and Matthew Willson and Alvaro Sanchez-Gonzalez and Peter Battaglia},
  year = {2025},
  note = {arXiv:2506.10772},
  url = {https://arxiv.org/abs/2506.10772}
}

@inproceedings{nguyen2025omnicast,
  title = {{OmniCast: A Masked Latent Diffusion Model for Weather Forecasting Across Time Scales}},
  author = {Tung Nguyen and Tuan Pham and Troy Arcomano and Veerabhadra Kotamarthi and Ian Foster and Sandeep Madireddy and Aditya Grover},
  booktitle = {Advances in Neural Information Processing Systems},
  volume = {38},
  year = {2025},
  url = {https://papers.nips.cc/paper_files/paper/2025/file/fb0b57dfb8686c357f96e41b880ce7ed-Paper-Conference.pdf}
}

@misc{sun2024llamagen,
  title = {{Autoregressive Model Beats Diffusion: Llama for Scalable Image Generation}},
  author = {Peize Sun and Yi Jiang and Shoufa Chen and Shilong Zhang and Bingyue Peng and Ping Luo and Zehuan Yuan},
  year = {2024},
  note = {arXiv:2406.06525},
  url = {https://arxiv.org/abs/2406.06525}
}

@inproceedings{chang2022maskgit,
  title = {{MaskGIT: Masked Generative Image Transformer}},
  author = {Huiwen Chang and Han Zhang and Lu Jiang and Ce Liu and William T. Freeman},
  booktitle = {IEEE/CVF Conference on Computer Vision and Pattern Recognition},
  year = {2022},
  pages = {11315--11325},
  url = {https://openaccess.thecvf.com/content/CVPR2022/html/Chang_MaskGIT_Masked_Generative_Image_Transformer_CVPR_2022_paper.html}
}

@inproceedings{hou2025treerl,
  title = {{TreeRL: LLM Reinforcement Learning with On-Policy Tree Search}},
  author = {Zhenyu Hou and Ziniu Hu and Yujiang Li and Rui Lu and Jie Tang and Yuxiao Dong},
  booktitle = {Annual Meeting of the Association for Computational Linguistics},
  year = {2025},
  pages = {12355--12369},
  doi = {10.18653/v1/2025.acl-long.604},
  url = {https://aclanthology.org/2025.acl-long.604/}
}

@inproceedings{li2026treepo,
  title = {{TreePO: Enhancing Policy Efficacy and Inference Efficiency with Tree Modeling}},
  author = {Yizhi Li and Qingshui Gu and Zhoufutu Wen and Ziniu Li and Ruibin Yuan and Tianshun Xing and Shuyue Guo and Tuney Zheng and Xin Zhou and Xingwei Qu and Wangchunshu Zhou and Zheng Zhang and Wei Shen and Wei Xue and Qian Liu and Chenghua Lin and Jian Yang and Ge Zhang and Wenhao Huang},
  booktitle = {International Conference on Machine Learning},
  year = {2026},
  volume = {306},
  pages = {66963--66976},
  url = {https://proceedings.mlr.press/v306/li26au.html}
}

@inproceedings{black2024ddpo,
  title = {{Training Diffusion Models with Reinforcement Learning}},
  author = {Kevin Black and Michael Janner and Yilun Du and Ilya Kostrikov and Sergey Levine},
  booktitle = {International Conference on Learning Representations},
  year = {2024},
  url = {https://arxiv.org/abs/2305.13301}
}

@inproceedings{wallace2024diffusiondpo,
  title = {{Diffusion Model Alignment Using Direct Preference Optimization}},
  author = {Bram Wallace and Meihua Dang and Rafael Rafailov and Linqi Zhou and Aaron Lou and Senthil Purushwalkam and Stefano Ermon and Caiming Xiong and Shafiq Joty and Nikhil Naik},
  booktitle = {IEEE/CVF Conference on Computer Vision and Pattern Recognition},
  year = {2024},
  pages = {8228--8238},
  url = {https://openaccess.thecvf.com/content/CVPR2024/html/Wallace_Diffusion_Model_Alignment_Using_Direct_Preference_Optimization_CVPR_2024_paper.html}
}

@inproceedings{liu2025flowgrpo,
  title = {{Flow-GRPO: Training Flow Matching Models via Online RL}},
  author = {Jie Liu and Gongye Liu and Jiajun Liang and Yangguang Li and Jiaheng Liu and Xintao Wang and Pengfei Wan and Di Zhang and Wanli Ouyang},
  booktitle = {Advances in Neural Information Processing Systems},
  volume = {38},
  year = {2025},
  url = {https://proceedings.neurips.cc/paper_files/paper/2025/hash/3a10c46572628d58cb44fb705f25cbbf-Abstract-Conference.html}
}

@misc{wang2025simplear,
  title = {{SimpleAR: Pushing the Frontier of Autoregressive Visual Generation through Pretraining, SFT, and RL}},
  author = {Junke Wang and Zhi Tian and Xun Wang and Xinyu Zhang and Weilin Huang and Zuxuan Wu and Yu-Gang Jiang},
  year = {2025},
  note = {arXiv:2504.11455},
  url = {https://arxiv.org/abs/2504.11455}
}

@misc{yuan2025argrpo,
  title = {{AR-GRPO: Training Autoregressive Image Generation Models via Reinforcement Learning}},
  author = {Shihao Yuan and Yahui Liu and Yang Yue and Jingyuan Zhang and Wangmeng Zuo and Qi Wang and Fuzheng Zhang and Guorui Zhou},
  year = {2025},
  note = {arXiv:2508.06924},
  url = {https://arxiv.org/abs/2508.06924}
}

@inproceedings{jiang2025t2ir1,
  title = {{T2I-R1: Reinforcing Image Generation with Collaborative Semantic-level and Token-level CoT}},
  author = {Dongzhi Jiang and Ziyu Guo and Renrui Zhang and Zhuofan Zong and Hao Li and Le Zhuo and Shilin Yan and Pheng-Ann Heng and Hongsheng Li},
  booktitle = {Advances in Neural Information Processing Systems},
  volume = {38},
  year = {2025},
  url = {https://papers.neurips.cc/paper_files/paper/2025/file/38fc6254f73450813db3b3e04397a9fc-Paper-Conference.pdf}
}

@misc{xu2025syncast,
  title = {{SynCast: Synergizing Contradictions in Precipitation Nowcasting via Diffusion Sequential Preference Optimization}},
  author = {Kaiyi Xu and Junchao Gong and Wenlong Zhang and Ben Fei and Lei Bai and Wanli Ouyang},
  year = {2025},
  note = {arXiv:2510.21847},
  url = {https://arxiv.org/abs/2510.21847}
}

@inproceedings{kang2026fakerair,
  title = {{Real-Time Long Horizon Air Quality Forecasting via Group-Relative Policy Optimization}},
  author = {Inha Kang and Eunki Kim and Wonjeong Ryu and Jaeyo Shin and Seungjun Yu and Yoon-Hee Kang and Seongeun Jeong and Eunhye Kim and Soontae Kim and Hyunjung Shim},
  booktitle = {IEEE/CVF Conference on Computer Vision and Pattern Recognition},
  year = {2026},
  url = {https://openaccess.thecvf.com/content/CVPR2026/papers/Kang_Real-Time_Long_Horizon_Air_Quality_Forecasting_via_Group-Relative_Policy_Optimization_CVPR_2026_paper.pdf}
}
}
\clearpage
\section*{Supplementary Material}
\suppressfloats[t]
\addcontentsline{toc}{section}{Supplementary Material}
\appendix

% Unified EB Garamond + soft purple/slate table palette matching main-text Tables 1 & 2
\definecolor{appink}{HTML}{292C42}
\definecolor{apppurple}{HTML}{544178}
\definecolor{appheader}{HTML}{E9E8F2}
\definecolor{appstripe}{HTML}{F6F7FA}
\definecolor{appours}{HTML}{EEE7F7}
\definecolor{apprankone}{HTML}{E4D9F1}
\definecolor{appranktwo}{HTML}{F7E7D6}
\newcommand{\appfont}{\fontfamily{EBGaramond-LF}\fontsize{9.0}{10.8}\selectfont\color{appink}}
\newcommand{\appsubhead}[2]{%
  \addlinespace[1.5pt]
  \multicolumn{#1}{@{}l}{\rule{0pt}{10.2pt}\textcolor{apppurple}{\bfseries\itshape #2}}\\[0.5pt]}

\section{Model Architecture and Pretraining}
\label{app:architecture}

\subsection{Notation and Tensor Organization}
\label{app:a-notation}
Table~\ref{tab:a-notation} collects the main notation used throughout the paper and appendix; batch dimensions are omitted. Each daily state contains $C=69$ standardized channels on a $120\times240$ grid, obtained by retaining the first 120 latitude rows of the native $121\times240$ grid. The tokenizer processes each day independently.

\begin{table*}[t]
\centering
\caption{\textbf{Main notation} shared by the main paper and appendix. Tensor shapes omit the batch dimension.}
\label{tab:a-notation}
\vspace{2pt}
\begingroup
\appfont
\setlength{\tabcolsep}{4pt}
\renewcommand{\arraystretch}{1.08}
\setlength{\aboverulesep}{1.8pt}
\setlength{\belowrulesep}{1.8pt}
\begin{tabularx}{\textwidth}{@{}p{0.18\textwidth}p{0.44\textwidth}>{\raggedright\arraybackslash}X@{}}
\toprule[0.95pt]
\rowcolor{appheader}
\rule{0pt}{10.5pt}\textbf{Symbol} & \textbf{Meaning} & \textbf{Shape or interpretation} \\
\midrule[0.55pt]
\rowcolor{appstripe} $X_t,\widehat X_t,\widetilde X_t$ & Observed, forecast, and reconstructed weather states & $C\times H\times W=69\times120\times240$ \\
$\mathcal F_\Theta,\pi_\theta$ & Weather predictive distribution and token dynamics policy & Field space; token space \\
\rowcolor{appstripe} $\mathcal E,\mathcal D,\mathcal T$ & Encoder, decoder, and discrete tokenizer & Weather--latent--token mappings \\
$f,\widehat f^s,z_t$ & Continuous, accumulated, and complete daily latents & $24\times60\times120$ \\
\rowcolor{appstripe} $r^s,r^{<s}$ & Scale-$s$ token map and its coarser prefix & $G\times H^s\times W^s$ indices \\
$(H^s,W^s)$ & Token grids, ordered from coarse to fine & $(15,30),(30,60),(60,120)$ \\
\rowcolor{appstripe} $S,G,D$ & Scales, FSQ groups, and axes per group & $3,4,6$ \\
$L_d,\mathcal Q$ & FSQ axis levels and index quantizer & $(L_1,\ldots,L_6)=(8,8,8,5,5,5)$ \\
\rowcolor{appstripe} $q,q^A,q^B$ & Composite index and its two categorical factors & $64{,}000=512\times125$ codes \\
$N,T$ & Composite tokens per day and forecast horizon & $N=37{,}800$; $T$ in days \\
\rowcolor{appstripe} $c_{t+1},h_t$ & Target annual phase and two-day forecast context & $2$; $(z_{t-1},z_t,c_{t+1})$ \\
$\mathsf H,\mathsf U^s,\mathsf V^s$ & Observed history, target query, and target content blocks & $14{,}400$, $H^sW^s$, $H^sW^s$ positions \\
\rowcolor{appstripe} $Y_\ell,M$ & Layer-$\ell$ hidden sequence and block-causal mask & Hidden width $1{,}536$ \\
$\mathcal L_{\mathrm{tok}},\mathcal L_{\mathrm{dyn}}$ & Tokenizer and dynamics pretraining objectives & Weighted MSE; token negative log likelihood \\
\rowcolor{appstripe} $\|\cdot\|_w^2$ & Latitude-weighted mean squared error & Channel-wise standardized space \\
$\mathcal S,\mathcal B$ & Search operator and sampled rollout record & Trajectories with shared prefixes \\
\rowcolor{appstripe} $K,\tau_i,R_i$ & Number of terminal trajectories, trajectory, and reward & $R_i=R(\tau_i)$ \\
$\pi_{\mathrm{old}},\pi_{\mathrm{ref}}$ & Rollout policy and reference policy & Frozen during a policy update \\
\rowcolor{appstripe} $A_v,A_e$ & Segment advantage and inherited token advantage & Sibling-relative reward normalization \\
$\rho_e,w_e,k_e$ & Token probability ratio, weight, and reference penalty & Terms in the clipped RL objective \\
\rowcolor{appstripe} $\epsilon,\beta,\mathcal J$ & Clipping radius, penalty coefficient, and RL objective & Policy optimization \\
\bottomrule[0.95pt]
\end{tabularx}
\endgroup
\end{table*}

\subsection{Multiscale Weather Tokenizer}
\label{app:a-tokenizer}
The tokenizer maps each daily field through a U-shaped convolutional encoder--decoder and a quantized bottleneck. Let $\operatorname{Conv}_{k}$ denote a $k\times k$ convolution, and GN denote GroupNorm. A residual block is
\begin{equation}
\begin{aligned}
v &= \operatorname{Conv}_{3}(\operatorname{SiLU}(\operatorname{GN}(x))),\\
\operatorname{Res}(x) &= x+\operatorname{Conv}_{3}(\operatorname{SiLU}(\operatorname{GN}(v))).
\end{aligned}
\label{eq:a-resblock}
\end{equation}
Convolutions use circular longitude padding and reflecting latitude padding. The encoder and decoder stages follow
\begin{equation}
\begin{aligned}
\operatorname{Enc}_j(x)
 &= \operatorname{Pool}_{a_j}\!\left(\operatorname{Conv}_{3}(\operatorname{ResStack}_{2}(x))\right),\\
\operatorname{Dec}_j(x)
 &= \operatorname{Conv}_{3}\!\left(\operatorname{Up}_{a_j}(\operatorname{ResStack}_{2}(x))\right),
\end{aligned}
\label{eq:a-stages}
\end{equation}
where $\operatorname{ResStack}_{k}$ denotes $k$ successive residual blocks with separate weights, Pool is average pooling, and Up is nearest-neighbor upsampling. Encoder factors are $a_j\in\{(1,2),(2,1),(1,1)\}$ in order; the decoder reverses them. After a 384-channel stem, encoder stages have widths $(512,768,768)$ and spatial sizes $(120,120)$, $(60,120)$, and $(60,120)$. Five residual blocks and a $1\times1$ projection yield $f\in\mathbb R^{24\times60\times120}$. The decoder applies the reverse bottleneck projection, five residual blocks, and stages of widths $(768,512,384)$, followed by GN, SiLU, and a 69-channel output convolution. All encoder information reaches the decoder through the quantized latent.

Residual FSQ~\cite{mentzer2024fsq} uses three grids (Table~\ref{tab:a-notation}), each with $G=4$ groups of $D=6$ axes and levels $(8,8,8,5,5,5)$. Area downsampling and bicubic upsampling implement the residual hierarchy in the main text, with unit scale gains. Each day contains $N=G\sum_s H^sW^s=37{,}800$ composite indices; the four groups share spatial positions in the Transformer.

\subsection{Autoregressive Dynamics Model}
\label{app:a-dynamics}
\paragraph{Context and scale ordering.}
Both daily latents enter $h_t=(z_{t-1},z_t,c_{t+1})$. Let $e_{\mathrm{row}}$, $e_{\mathrm{col}}$, $e^s$, and $e_\rho$ be learned row, column, scale, and role embeddings, respectively. The role $\rho$ distinguishes the two history days, target queries, and target content. At each spatial position, we embed a pair of latent vectors as
\begin{equation}
\begin{aligned}
\operatorname{Emb}(a,b;s,\rho)
 &= \operatorname{Linear}([a;b])+e_{\mathrm{row}}+e_{\mathrm{col}}\\
 &\quad +e^s+e_\rho+\operatorname{MLP}(c_{t+1}),
\end{aligned}
\label{eq:a-embedding}
\end{equation}
Here $[a;b]$ concatenates channels; all embeddings have width 1,536, and the annual MLP uses SiLU. The observed block $\mathsf H$ concatenates $\operatorname{Emb}(z_{t-1},0;S,\rho_{t-1})$ and $\operatorname{Emb}(z_t,0;S,\rho_t)$ over 14,400 spatial positions. At scale $s$, query block $\mathsf U^s$ embeds the resized pair $(z_t,\widehat f_{t+1}^{s-1})$; content block $\mathsf V^s$ embeds $(z_t,\widehat f_{t+1}^{s})$. The flattened sequence is
\begin{equation}
Y_0=[\mathsf H;\mathsf U^1;\mathsf V^1;\mathsf U^2;\mathsf V^2;\mathsf U^3].
\label{eq:a-sequence}
\end{equation}
Attention is bidirectional within each block and causal across blocks. Thus, queries access both history days and strictly coarser target content. Content uses ground-truth tokens during training and sampled tokens during inference.

\paragraph{Transformer and output heads.}
The 16 Transformer blocks use hidden width 1,536, 24 attention heads, and a GELU feed-forward network (FFN) of width 6,144:
\begin{equation}
\begin{aligned}
\widetilde Y_\ell &= Y_{\ell-1}+\operatorname{DP}_\ell\!\left(
 \operatorname{MHA}_{M}(\operatorname{LN}(Y_{\ell-1}))\right),\\
Y_\ell &= \widetilde Y_\ell+\operatorname{DP}_\ell\!\left(
 \operatorname{FFN}(\operatorname{LN}(\widetilde Y_\ell))\right).
\end{aligned}
\label{eq:a-transformer}
\end{equation}
Here LN is LayerNorm, $M$ is the block-causal mask, and DP is stochastic depth with drop rates increasing from 0 to 0.1. A final LN produces the query features $y$ used for prediction.

Following MAGVIT-v2's vocabulary factorization~\cite{yu2024lfq}, two linear heads produce $G\times512$ and $G\times125$ logits per position. For each group, the composite index and its conditional probability satisfy
\begin{equation}
\begin{aligned}
q^A &= q\bmod512,\qquad q^B=\lfloor q/512\rfloor,\\
q &= q^A+512q^B,\\
p(q\mid y) &= p_\theta^{(A)}(q^A\mid y)\,p_\theta^{(B)}(q^B\mid y),
\end{aligned}
\label{eq:a-factorization}
\end{equation}
where $q\in\{0,\ldots,63999\}$ and each factor distribution is a head softmax. The two factor log probabilities add in likelihood training and policy optimization. Spatial positions and groups are sampled in parallel within each scale.

\paragraph{Forecast rollout.}
For each layer and attention head, let $Q_i,K_i,V_i$ be the projections of the current block in Eq.~\eqref{eq:a-sequence}. With layer and head indices omitted, cached attention is
\begin{equation}
\begin{aligned}
K_{\leq i} &= [K_{<i};K_i],\quad V_{\leq i}=[V_{<i};V_i],\\
O_i &= \operatorname{softmax}\!\left(
 Q_iK_{\leq i}^{\mathsf T}/\sqrt{d_h}\right)V_{\leq i},
\end{aligned}
\label{eq:a-kv-cache}
\end{equation}
where concatenation is along positions and $d_h=64$. The cache retains keys and values from the history and completed query/content blocks, so each new block reuses prior projections while attending to its entire current block. Completing a day advances the context to $(z_t,z_{t+1},c_{t+2})$ using the generated latent directly. Caches are rebuilt for the new history and target date.

\subsection{Pretraining Configuration}
\label{app:a-training}
The tokenizer minimizes latitude-weighted reconstruction MSE in standardized space. With the tokenizer frozen, dynamics pretraining minimizes next-day token negative log likelihood conditioned on two days, summing both factor losses and averaging over composite tokens and examples.

Both models are trained for 50 epochs with AdamW, linear learning-rate warmup, and cosine decay. Table~\ref{tab:a-training} lists the optimization settings. We use bfloat16 mixed precision and clip the gradient norm to 1.0. Dynamics training additionally uses activation checkpointing. Checkpoints are selected by validation reconstruction error for the tokenizer and validation joint token negative log likelihood for dynamics.

\begin{table}[t]
\centering
\caption{\textbf{Pretraining configuration.} Batch sizes are global across devices.}
\label{tab:a-training}
\vspace{2pt}
\begingroup
\appfont
\setlength{\tabcolsep}{4pt}
\renewcommand{\arraystretch}{1.12}
\setlength{\aboverulesep}{1.8pt}
\setlength{\belowrulesep}{1.8pt}
\begin{tabularx}{\linewidth}{@{}Xcc@{}}
\toprule[0.95pt]
\rowcolor{appheader}
\rule{0pt}{10.5pt}\textbf{Setting} & \textbf{Tokenizer} & \textbf{Dynamics} \\
\midrule[0.55pt]
\rowcolor{appstripe} Optimizer & AdamW & AdamW \\
Epochs & 50 & 50 \\
\rowcolor{appstripe} Global batch size & 32 & 64 \\
Peak learning rate & $2\times10^{-4}$ & $2\times10^{-4}$ \\
\rowcolor{appstripe} Minimum learning rate & $10^{-5}$ & $2\times10^{-5}$ \\
Warmup updates & 200 & 500 \\
\rowcolor{appstripe} Weight decay & 0.01 & 0.05 \\
Gradient norm limit & 1.0 & 1.0 \\
\rowcolor{appstripe} Precision & bfloat16 & bfloat16 \\
\bottomrule[0.95pt]
\end{tabularx}
\endgroup
\end{table}

\section{Weather Tokenization}
\label{app:tokenization}
\subsection{Reconstruction Quality and Group Capacity}
\label{app:b-reconstruction}
\paragraph{Setup.}
We evaluate residual FSQ tokenizers with $G\in\{2,4,8\}$ parallel groups, using the ERA5 data split and channel standardization in Sec.~\ref{sec:exp-setup}. The encoder--decoder backbone, three spatial grids, and six-axis FSQ vocabulary follow Sec.~\ref{app:a-tokenizer}; varying $G$ changes the bottleneck width to $6G$ and the daily token count to $9{,}450G$. We report latitude-weighted reconstruction nRMSE on the 2019 validation set and the 2020--2022 test set. The controlled group comparison uses a common 30-epoch budget and validation-selected checkpoints. The final four-group tokenizer is also reported at its 50-epoch training budget.

\begin{table*}[t]
\centering
\caption{\textbf{FSQ group capacity.} Reconstruction nRMSE ($\downarrow$). The first three rows use a common $30$-epoch budget; the highlighted final row is the $50$-epoch tokenizer used by \weathercraft{}. Q700 and V850 report test errors for specific humidity at $700$\,hPa and meridional wind at $850$\,hPa.}
\label{tab:tokenizer_groups}
\vspace{2pt}
\begingroup
\appfont
\setlength{\tabcolsep}{4pt}
\renewcommand{\arraystretch}{1.12}
\setlength{\aboverulesep}{1.8pt}
\setlength{\belowrulesep}{1.8pt}
\begin{tabularx}{\textwidth}{@{}*{7}{>{\centering\arraybackslash}X}@{}}
\toprule[0.95pt]
\rowcolor{appheader}
\rule{0pt}{10.5pt}\textbf{Groups $G$} & \textbf{Epochs} & \textbf{Tokens/day} & \textbf{Validation} & \textbf{Test} & \textbf{Q700} & \textbf{V850} \\
\midrule[0.55pt]
\rowcolor{appstripe} 2 & 30 & 18,900 & 0.0935 & 0.0926 & 0.1406 & 0.1288 \\
4 & 30 & 37,800 & 0.0698 & 0.0693 & 0.0997 & 0.1015 \\
\rowcolor{appstripe} 8 & 30 & 75,600 & 0.0667 & 0.0662 & 0.0956 & 0.0941 \\
\midrule[0.55pt]
\rowcolor{appours}
\rule{0pt}{11.0pt}\textcolor{apppurple}{\bfseries 4 (ours)} & \textbf{50} & \textbf{37,800} & \textbf{0.0669} & \textbf{0.0664} & \textbf{0.0957} & \textbf{0.0972} \\
\bottomrule[0.95pt]
\end{tabularx}
\endgroup
\end{table*}

\paragraph{Effect of group capacity.}
Increasing $G$ from two to four reduces test reconstruction nRMSE from 0.0926 to 0.0693, a 25.2\% improvement at the same training budget (Table~\ref{tab:tokenizer_groups}). Doubling the group count again reduces nRMSE to 0.0662, a further 4.5\% improvement. The additional capacity is particularly useful for moisture and wind: four groups reduce Q700 and V850 reconstruction errors by 29.1\% and 21.2\%, respectively, relative to two groups. With the full pretraining budget, our four-group tokenizer reaches a test nRMSE of 0.0664.

\begin{figure*}[t]
\centering
\includegraphics[width=\textwidth]{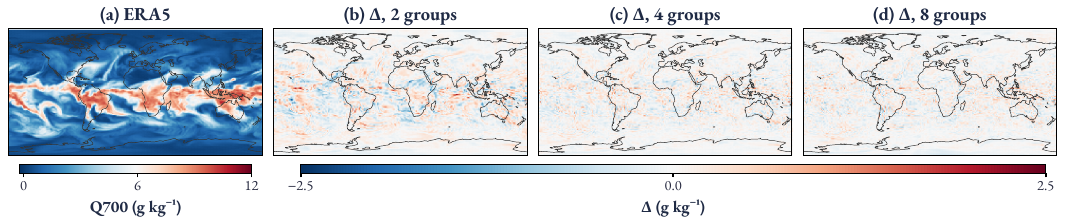}
\caption{\textbf{Q700 reconstruction.} ERA5 700-hPa specific humidity on 4 January 2020 and reconstruction errors ($\Delta=\widetilde X-X$) for two-, four-, and eight-group tokenizers trained for 30 epochs. Error panels share a symmetric color scale.}
\label{fig:tokenizer-q-700}
\end{figure*}

\begin{figure*}[t]
\centering
\includegraphics[width=\textwidth]{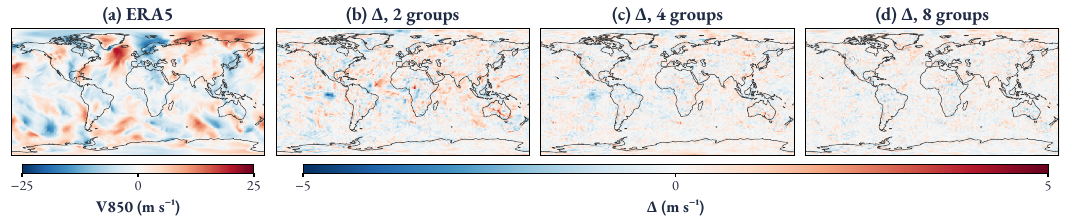}
\caption{\textbf{V850 reconstruction.} ERA5 850-hPa meridional wind on 4 January 2020 and reconstruction errors ($\Delta=\widetilde X-X$) for two-, four-, and eight-group tokenizers trained for 30 epochs. Error panels share a symmetric color scale.}
\label{fig:tokenizer-v-850}
\end{figure*}

\paragraph{Spatial reconstruction.}
Figures~\ref{fig:tokenizer-q-700} and~\ref{fig:tokenizer-v-850} show the ERA5 fields and signed reconstruction errors for Q700 and V850. Shared error scales expose the spatial effect of increasing group capacity in the controlled comparison.

\paragraph{Choice for weather dynamics.}
We adopt four groups to combine accurate reconstruction with a compact representation for autoregressive modeling. This configuration uses 24 latent channels and 37,800 composite tokens per day, half the corresponding eight-group budgets. Groups share spatial positions, so their number controls the input-projection and categorical-output dimensions while the Transformer sequence length stays fixed. Four groups retain most of the reconstruction improvement in the controlled comparison and reduce the number of categorical predictions required for each forecast.

\paragraph{Compression budget.}
Across all scales and groups, the 37,800 composite codes require
$37{,}800\times16=604{,}800$ bits per daily state when packed into
16-bit indices. Relative to the $69\times120\times240$ FP32 input,
this gives $105.1\times$ compression, or 0.304 bits per input scalar,
excluding file metadata and without entropy coding. This compact
representation supports both weather reconstruction and autoregressive
probabilistic forecasting. Related scientific tokenizers and atmospheric
compressors explore complementary choices of variable grouping,
quantization, and coding~\cite{lingsch2026phaedra,mirowski2024neuralcompressionatmosphericstates,zhao2025transformingweatherdatapixel}.
Our group-capacity study selects a representation budget for the complete
forecasting pipeline. Future work will explore more efficient discrete
compression architectures and allocation of token capacity across
variables and spatial scales.

\section{Post-Training Algorithms and Downstream Tasks}
\label{app:post-training}
We adapt the pretrained dynamics policy through two branches: direct optimization of probabilistic scores and reinforcement learning with event rewards. Both retain the frozen tokenizer and decoder from Sec.~\ref{app:a-tokenizer}.

\subsection{CRPS Fine-Tuning}
\label{app:c-crps}
\paragraph{Objective and gradient.}
We generate two independent trajectories from $h_t$ and score their terminal fields at lead $T$. For standardized fields, let $d_w(U,V)$ be the absolute error averaged over channels and grid points with normalized cosine-latitude weights. The two-member fair-CRPS objective~\cite{ferro2014fairscores} is
\begin{equation}
\begin{aligned}
\mathcal L_{\mathrm{CRPS}}^{(T)}=\tfrac12\big[&d_w(\widehat X_{t+T}^{(1)},X_{t+T})\\
 &+d_w(\widehat X_{t+T}^{(2)},X_{t+T})\\
 &-d_w(\widehat X_{t+T}^{(1)},\widehat X_{t+T}^{(2)})\big].
\end{aligned}
\label{eq:c-crps}
\end{equation}
For each categorical head with logits $\lambda$, independent standard Gumbel noise $g$ gives the straight-through sample
\begin{equation}
\begin{aligned}
a&=\operatorname{onehot}(\arg\max(\lambda+g)),\\
p_{\tau}&=\operatorname{softmax}((\lambda+g)/\tau),\\
\widetilde a&=a+p_{\tau}-\operatorname{sg}(p_{\tau}),\qquad \tau=1.
\end{aligned}
\label{eq:c-ste}
\end{equation}
Here $\operatorname{sg}$ stops gradients. Multiplication by each factor's fixed FSQ lookup vectors yields a hard code in the forward pass and a relaxed gradient in the backward pass. Gradients pass through the frozen decoder inputs and through all generated scales and days. This is a biased straight-through estimator of the marginal fair-CRPS objective. Algorithm~\ref{alg:c-crps} summarizes the update.

\begin{algorithm}[t]
\caption{CRPS fine-tuning}
\label{alg:c-crps}
\begin{algorithmic}[1]
\Require Pretrained $\pi_\theta$; frozen $\mathcal T,\mathcal D$
\For{$T=1,2$ in order}
  \State Initialize AdamW for the current stage
  \For{each update}
    \State Sample a batch of observed histories and targets
    \State Encode each history as $h_t=(z_{t-1},z_t,c_{t+1})$
    \State Generate two independent $T$-day trajectories using Eq.~\eqref{eq:c-ste}
    \State Decode their terminal latents with $\mathcal D$
    \State Average Eq.~\eqref{eq:c-crps} over the batch
    \State Backpropagate through all scales and days; update $\theta$
  \EndFor
  \State Retain the checkpoint with lowest validation fair-CRPS at lead $T$
\EndFor
\end{algorithmic}
\end{algorithm}

\paragraph{Curriculum and evaluation.}
Following the short-to-long terminal-score curriculum used by Tyche~\cite{xu2026tyche}, we first optimize $T=1$ ($1$-day CRPS) and then $T=2$ ($2$-day CRPS), at daily resolution. The $1$-day stage uses 500 updates; the $2$-day continuation uses a 750-update budget and selects update 625 on 2019 validation forecasts. Each stage uses AdamW with learning rate $10^{-6}$, $(\beta_1,\beta_2)=(0.9,0.999)$, zero weight decay, global batch size 32, and gradient norm clipping at 1. Sampling uses temperature one and the full vocabulary. Dynamics forward passes use bfloat16, with float32 code lookup and decoding; activation checkpointing preserves the complete rollout gradient.

Table~\ref{tab:c-crps} complements the main fair-CRPS comparison with ensemble-mean nRMSE on the same eight-member test forecasts. The two-day stage improves nRMSE across all reported leads and extends the fair-CRPS gains to longer forecasts.
\begin{table*}[t]
\centering
\caption{\textbf{CRPS adaptation.} Eight-member forecasts on the main test set. Bold and underlining indicate the best and second-best values within each metric, ranked before rounding.}
\label{tab:c-crps}
\vspace{2pt}
\begingroup
\appfont
\setlength{\tabcolsep}{4pt}
\renewcommand{\arraystretch}{1.12}
\setlength{\aboverulesep}{1.8pt}
\setlength{\belowrulesep}{1.8pt}
\begin{tabularx}{\textwidth}{@{}p{0.24\textwidth}*{5}{>{\centering\arraybackslash}X}@{}}
\toprule[0.95pt]
\rowcolor{appheader}
\rule{0pt}{10.5pt}\textbf{Model} & \textbf{1-Day} & \textbf{3-Day} & \textbf{5-Day} & \textbf{10-Day} & \textbf{15-Day} \\
\midrule[0.55pt]
\appsubhead{6}{Ensemble-mean nRMSE $\downarrow$}
\rowcolor{appstripe} Base & 0.1453 & 0.2586 & 0.3721 & 0.5474 & 0.5967 \\
+ $1$-day CRPS & \underline{0.1449} & \underline{0.2553} & \underline{0.3654} & \underline{0.5413} & \underline{0.5930} \\
\rowcolor{appours}
\rule{0pt}{11.0pt}\textcolor{apppurple}{\bfseries + $2$-day CRPS} & \textbf{0.1446} & \textbf{0.2530} & \textbf{0.3618} & \textbf{0.5385} & \textbf{0.5906} \\
\midrule[0.55pt]
\appsubhead{6}{Fair-CRPS $\downarrow$}
\rowcolor{appstripe} Base & 0.0602 & 0.1065 & 0.1526 & 0.2252 & 0.2476 \\
+ $1$-day CRPS & \textbf{0.0582} & \underline{0.1017} & \underline{0.1457} & \underline{0.2197} & \underline{0.2436} \\
\rowcolor{appours}
\rule{0pt}{11.0pt}\textcolor{apppurple}{\bfseries + $2$-day CRPS} & \underline{0.0584} & \textbf{0.1008} & \textbf{0.1441} & \textbf{0.2181} & \textbf{0.2421} \\
\bottomrule[0.95pt]
\end{tabularx}
\endgroup
\end{table*}

\subsection{Vanilla GRPO and Tree GRPO}
\label{app:c-grpo}
Both methods optimize the clipped objective in Eq.~\eqref{eq:method-12}, using composite-token log probabilities summed over the two categorical heads. The reference policy $\pi_{\mathrm{ref}}$ remains fixed at the pretrained model; $\pi_{\mathrm{old}}$ is refreshed before each rollout round. For a sampled token occurrence $e$, the reference penalty is
\begin{equation}
\begin{aligned}
\delta_e&=\log\pi_{\mathrm{ref}}(a_e\mid h_e)-\log\pi_\theta(a_e\mid h_e),\\
k_e&=\exp(\delta_e)-\delta_e-1.
\end{aligned}
\label{eq:c-kl}
\end{equation}
Rollout log probabilities, rewards, and advantages are held fixed during the two updates that reuse a rollout batch. Sampling and likelihood replay disable stochastic depth so they evaluate the same conditional policy.

\paragraph{Advantages and aggregation.}
Vanilla GRPO~\cite{shao2024grpo} samples $K=8$ independent trajectories, assigning all tokens in trajectory $i$ the advantage $A_i=(R_i-\overline R)/\sigma_R$. Tree GRPO branches at days 1, 4, and 7 with widths $2\rightarrow4\rightarrow8$. Each newly generated segment receives the descendant-mean, sibling-relative advantage in Eq.~\eqref{eq:method-9}. Both use the population standard deviation of the eight terminal rewards, with zero advantages when $\sigma_R<10^{-4}$.

Let $n_s=GH^sW^s$ and $B_\ell$ be the number of active branches at lead $\ell$. The normalized weight of a token at that lead and scale is
\begin{equation}
\begin{aligned}
w_e&=\frac{\alpha_s}{T B_\ell n_s},\\
\alpha_s&=\begin{cases}
n_s/N, & \text{Vanilla},\\
\sqrt{n_s}/\sum_j\sqrt{n_j}, & \text{Tree}.
\end{cases}
\end{aligned}
\label{eq:c-weights}
\end{equation}
Vanilla uses $B_\ell=8$ throughout; Tree uses $2$, $4$, and $8$ over leads 1--3, 4--6, and 7--$T$. Their scale totals are therefore $1{:}4{:}16$ and $1{:}2{:}4$, respectively. Tree stores each shared prefix once and assigns credit only to newly generated tokens. Objectives are additionally averaged over the batch of origins.

\begin{algorithm}[t]
\caption{Vanilla GRPO}
\label{alg:c-vanilla}
\begin{algorithmic}[1]
\Require Pretrained $\pi_\theta$, frozen $\pi_{\mathrm{ref}}$, task reward $R$
\For{each rollout round}
  \State Set $\pi_{\mathrm{old}}\gets\pi_\theta$; sample a batch of contexts
  \State Sample $K=8$ independent trajectories per context
  \State Evaluate $R_i=R(\tau_i)$ and store sampled-token log probabilities
  \State Compute group-normalized $A_i$; assign it to trajectory $i$'s tokens
  \For{two updates on this rollout batch}
    \State Replay sampled tokens under $\pi_\theta$
    \State Update $\theta$ by Eq.~\eqref{eq:method-12} with Vanilla weights in Eq.~\eqref{eq:c-weights}
  \EndFor
\EndFor
\end{algorithmic}
\end{algorithm}

\begin{algorithm}[t]
\caption{Tree GRPO}
\label{alg:c-tree}
\begin{algorithmic}[1]
\Require Pretrained $\pi_\theta$, frozen $\pi_{\mathrm{ref}}$, task reward $R$
\For{each rollout round}
  \State Set $\pi_{\mathrm{old}}\gets\pi_\theta$; sample a batch of contexts
  \State Grow a rollout tree, branching at days 1, 4, and 7
  \State Score its eight leaves and compute their reward standard deviation
  \State Assign each new segment its sibling-relative advantage by Eq.~\eqref{eq:method-9}
  \State Store unique tokens, inherited advantages, and rollout log probabilities
  \For{two updates on this rollout batch}
    \State Replay each unique segment under $\pi_\theta$
    \State Update $\theta$ by Eq.~\eqref{eq:method-12} with Tree weights in Eq.~\eqref{eq:c-weights}
  \EndFor
\EndFor
\end{algorithmic}
\end{algorithm}

\subsection{Downstream Rewards and Experimental Settings}
\label{app:c-downstream}
\paragraph{Calendar-dependent thresholds.}
For temperature $Y$ at grid cell $x$, the percentile threshold for calendar day $d$ is computed from reference years $\mathcal Y$ using an 11-day pool, followed by 31-day circular smoothing:
\begin{equation}
\begin{aligned}
\widetilde\vartheta_p(d,x)&=Q_p\!\left(\{Y(y,d',x):y\in\mathcal Y,\right.\\[-2pt]
 &\hspace{47pt}\left.\operatorname{dist}_{\mathrm{cyc}}(d',d)\leq5\}\right),\\
\vartheta_p(d,x)&=\frac1{31}\sum_{j=-15}^{15}\widetilde\vartheta_p(d+j,x).
\end{aligned}
\label{eq:c-threshold}
\end{equation}
Actual leap days enter the reference samples. Thresholds and masks are fixed before policy optimization.

\paragraph{Persistence and CSI.}
For a daily exceedance indicator $b_d(x)$, define $\operatorname{Persist}_k(b)_d(x)$ to mark every day belonging to a complete run of at least $k$ consecutive exceedances. Predicted and observed event maps are scored over the designated lead window $\mathcal W$ and spatial domain $\Omega$:
\begin{equation}
\operatorname{CSI}_{\mathcal W}
=\frac{\sum_{\ell\in\mathcal W,x\in\Omega}\omega_x\widehat E_{\ell,x}E_{\ell,x}}
{\sum_{\ell\in\mathcal W,x\in\Omega}\omega_x
\mathbf1[\widehat E_{\ell,x}\lor E_{\ell,x}]},
\quad \omega_x=\cos\varphi_x.
\label{eq:c-csi}
\end{equation}
This pools area-weighted hits, false alarms, and misses before division; an empty union has score zero. Persistence labeling uses four observed days followed by each sampled forecast trajectory; the weather policy itself retains its two-day context. Evaluation averages trajectory CSI over eight independent members and then over origins.

\paragraph{Land temperature events.}
We use ERA5 daily-mean T2M thresholds from 1981--2010 and the Natural Earth land mask between $60^\circ$S and $75^\circ$N. Warm and cold events are $\operatorname{Persist}_3(Y>\vartheta_{0.9})$ and $\operatorname{Persist}_3(Y<\vartheta_{0.1})$. We score the 15-day rollout over a fixed interior window, leads 3--11, retaining subsequent forecast days as context for persistence labeling. Warm and cold CSI are computed separately and averaged, giving each tail equal weight in the reward:
\begin{equation}
R_{\mathrm{land}}=\tfrac12\big(\operatorname{CSI}^{\mathrm{warm}}_{3:11}
+\operatorname{CSI}^{\mathrm{cold}}_{3:11}\big).
\label{eq:c-land-reward}
\end{equation}

\paragraph{Ocean coupling and SST events.}
Each sampled weather trajectory supplies physical daily-mean U10, V10, T2M, and MSLP to a frozen TritonCast~\cite{wu2025tritoncast}. Coordinate-aware bilinear interpolation maps the forcing to its $0.25^\circ$ ocean grid. The model advances its 93-channel ocean state from the observed GLORYS initial condition; its fixed normalization and daily mean climatology recover physical SST. All methods use the same wet-cell and finite-data mask, including threshold availability. The author's mean climatology is used for anomaly recovery; percentile thresholds are estimated separately from GLORYS SST over 1993--2018 using Eq.~\eqref{eq:c-threshold}.

The ocean reward uses daily q90 exceedances over a 10-day rollout:
\begin{equation}
R_{\mathrm{ocean}}=\operatorname{CSI}^{\mathrm{SST}>\vartheta_{0.9}}_{3:10}.
\label{eq:c-ocean-reward}
\end{equation}
We additionally evaluate q95 exceedances and persistent q90 events lasting at least five days. Persistent-event CSI uses leads 3--6, where complete duration context is available. A 1993--2012 threshold baseline provides a reference-period sensitivity check. Ocean threshold pools require at least 90\% of reference years, 95\% of expected daily samples, and a complete smoothing window. No gradients pass through TritonCast, thresholding, persistence detection, or CSI; their scalar rewards train the weather policy through sampled-token probabilities.

\paragraph{RL pipeline and settings.}
Land and ocean policies are initialized independently from the same pretrained dynamics model. Vanilla and Tree use the same 400 training origins, arranged into 50 rounds of eight origins with two updates per rollout batch, for 100 optimizer steps each. Training dates lie in 2013--2018; both methods also share the validation and test origins. Validation uses 16 origins in 2019, evaluated every eight updates and at the final update; the base model is included in checkpoint selection. The test set contains 128 origins across 2020--2022. Evaluation samples eight independent trajectories directly from the selected policy. Table~\ref{tab:c-settings} summarizes the shared optimization settings.
\begin{table}[t]
\centering
\caption{\textbf{GRPO settings.} Frozen components are the tokenizer, decoder, reference policy, and (for ocean tasks) TritonCast.}
\label{tab:c-settings}
\vspace{2pt}
\begingroup
\appfont
\setlength{\tabcolsep}{4pt}
\renewcommand{\arraystretch}{1.12}
\setlength{\aboverulesep}{1.8pt}
\setlength{\belowrulesep}{1.8pt}
\begin{tabularx}{\linewidth}{@{}X>{\centering\arraybackslash}p{0.34\linewidth}@{}}
\toprule[0.95pt]
\rowcolor{appheader}
\rule{0pt}{10.5pt}\textbf{Setting} & \textbf{Value} \\
\midrule[0.55pt]
\appsubhead{2}{Optimization}
\rowcolor{appstripe} Optimizer & AdamW \\
Learning rate & $10^{-6}$ \\
\rowcolor{appstripe} Weight decay & 0 \\
AdamW $(\beta_1,\beta_2)$ & $(0.9,0.95)$ \\
\rowcolor{appstripe} Gradient norm limit & 1.0 \\
Policy clip $\epsilon$ / KL weight $\beta$ & $0.2$ / $0.02$ \\
\midrule[0.55pt]
\appsubhead{2}{Rollout and updates}
\rowcolor{appstripe} Origins / round & 8 \\
Updates / round & 2 \\
\rowcolor{appstripe} Rounds / total updates & 50 / 100 \\
Rollout leaves / test members & 8 / 8 \\
\rowcolor{appstripe} Sampling temperature & 1 \\
Sampling vocabulary & Full \\
\rowcolor{appstripe} Rollout days (land / ocean) & 15 / 10 \\
\midrule[0.55pt]
\appsubhead{2}{Selected update (Vanilla / Tree)}
\rowcolor{appstripe} Land & 24 / 100 \\
Ocean & 72 / 100 \\
\bottomrule[0.95pt]
\end{tabularx}
\endgroup
\end{table}

\paragraph{Supplementary scores.}
Table~\ref{tab:c-events} reports event and continuous-field metrics. Ocean gap closure in the main text is computed from mean CSI values as $100(\overline{\mathrm{CSI}}_{\mathrm{RL}}-\overline{\mathrm{CSI}}_{\mathrm{base}})/(\overline{\mathrm{CSI}}_{\mathrm{ERA5}}-\overline{\mathrm{CSI}}_{\mathrm{base}})$. The ERA5-forcing control advances the same frozen ocean model from the same initial state. We estimate 95\% confidence intervals by resampling 30-calendar-day blocks of forecast initializations within each test year, repeating this procedure 10,000 times. Each replicate uses the same sampled blocks for all methods and recomputes the gain over the pretrained model. Resampling blocks preserves dependence among nearby initializations, while pairing the samples keeps the method comparisons on the same dates.
\begin{table*}[t]
\centering
\caption{\textbf{Downstream evaluation.} Methods are compared within each task; weather RMSE is channel-normalized. SST and T2M RMSE aggregate member squared errors over the scored event window. Bold and underlining mark the best and second-best values before rounding.}
\label{tab:c-events}
\vspace{2pt}
\begingroup
\appfont
\setlength{\tabcolsep}{4pt}
\renewcommand{\arraystretch}{1.12}
\setlength{\aboverulesep}{1.8pt}
\setlength{\belowrulesep}{1.8pt}
\begin{tabularx}{\textwidth}{@{}p{0.24\textwidth}*{5}{>{\centering\arraybackslash}X}@{}}
\toprule[0.95pt]
\rowcolor{appheader}
\rule{0pt}{10.5pt}\textcolor{apppurple}{\bfseries\itshape Ocean SST events} & \multicolumn{3}{c}{\textbf{CSI $\uparrow$}} & \multicolumn{2}{c@{}}{\textbf{RMSE $\downarrow$}} \\[-0.5pt]
\cmidrule(lr){2-4}\cmidrule(l){5-6}
\textbf{Method} & \textbf{q90} & \textbf{q95} & \textbf{$\ge 5$ days} & \textbf{SST ($^\circ$C)} & \textbf{Weather} \\
\midrule[0.55pt]
\rowcolor{appstripe} Base & 0.5561 & 0.4730 & 0.6448 & \underline{0.3849} & \textbf{0.5029} \\
Vanilla GRPO & \underline{0.5576} & \underline{0.4749} & \underline{0.6461} & \textbf{0.3848} & \underline{0.5054} \\
\rowcolor{appours}
\rule{0pt}{11.0pt}\textcolor{apppurple}{\bfseries Tree GRPO} & \textbf{0.5598} & \textbf{0.4776} & \textbf{0.6472} & 0.3864 & 0.5088 \\
\midrule[0.65pt]
\rowcolor{appheader}
\rule{0pt}{10.5pt}\textcolor{apppurple}{\bfseries\itshape Land temperature events} & \multicolumn{3}{c}{\textbf{CSI $\uparrow$}} & \multicolumn{2}{c@{}}{\textbf{RMSE $\downarrow$}} \\[-0.5pt]
\cmidrule(lr){2-4}\cmidrule(l){5-6}
\textbf{Method} & \textbf{Mean} & \textbf{Warm} & \textbf{Cold} & \textbf{T2M (K)} & \textbf{Weather} \\
\midrule[0.55pt]
\rowcolor{appstripe} Base & 0.1767 & 0.2208 & 0.1325 & \textbf{3.3437} & \textbf{0.6027} \\
Vanilla GRPO & \underline{0.1791} & \underline{0.2232} & \underline{0.1349} & \underline{3.3592} & \underline{0.6064} \\
\rowcolor{appours}
\rule{0pt}{11.0pt}\textcolor{apppurple}{\bfseries Tree GRPO} & \textbf{0.1822} & \textbf{0.2286} & \textbf{0.1359} & 3.3863 & 0.6088 \\
\bottomrule[0.95pt]
\end{tabularx}
\endgroup
\end{table*}

\section{Evaluation and Ablation Studies}
\label{app:eval-ablation}

\subsection{Evaluation Metrics}
\label{app:d-metrics}

\paragraph{Standardization and spatial weighting.}
Following the main text, $X_t\in\mathbb R^{C\times H\times W}$ denotes
a standardized weather state, where $C$ is the number of channels and
$H$ and $W$ are the numbers of latitude and longitude grid points.
For channel $v$, physical values are recovered as
$X^{\mathrm{phys}}_{t,v}=\mu_v+\sigma_v X_{t,v}$, where $\mu_v$ and
$\sigma_v$ are estimated from the training set. Let $i$ and $j$ index
latitude and longitude, and let $\varphi_i$ be latitude. For a scalar
spatial field $U\in\mathbb R^{H\times W}$, define its normalized
latitude-weighted spatial average as
\begin{equation}
\begin{aligned}
w_i&=\frac{\cos\varphi_i}{W\sum_{i'=1}^{H}\cos\varphi_{i'}},\\
\langle U\rangle_w&=\sum_{i=1}^{H}\sum_{j=1}^{W}w_i U_{ij}.
\end{aligned}
\label{eq:d-area}
\end{equation}
Here $\langle\cdot\rangle_w$ denotes a weighted average, including when
applied to pointwise products or squared errors below, and
$\sum_{i,j}w_i=1$. The shorthand in the tokenizer objective expands to
\begin{equation}
\|X-\widetilde X\|_w^2
=\frac{1}{C}\sum_{v=1}^{C}\sum_{i=1}^{H}\sum_{j=1}^{W}
w_i\big(X_{vij}-\widetilde X_{vij}\big)^2.
\label{eq:d-norm}
\end{equation}
Each channel contributes equally in standardized space.

\paragraph{Root mean squared error.}
Let $\mathcal I$ contain the forecast initialization dates, and let
$\widehat X^{(k)}_{t+\ell}$ denote member $k$ at lead $\ell$ from
initialization $t$. The point forecast is the ensemble mean
$\overline X_{t+\ell}=K^{-1}\sum_{k=1}^{K}\widehat X^{(k)}_{t+\ell}$;
for deterministic models, $K=1$. At a fixed lead, we compute
\begin{equation}
\begin{aligned}
e_{v,\ell}&=\frac{1}{|\mathcal I|}\sum_{t\in\mathcal I}
\left\langle(\overline X_{t+\ell,v}-X_{t+\ell,v})^2\right\rangle_w,\\
\operatorname{nRMSE}_{\ell}&=\sqrt{\frac{1}{C}\sum_{v=1}^{C}e_{v,\ell}},\\
\operatorname{RMSE}_{v,\ell}&=\sigma_v\sqrt{e_{v,\ell}}.
\end{aligned}
\label{eq:d-rmse}
\end{equation}
The last line gives channel-wise RMSE in physical units. Scores summarized
over multiple leads average squared errors over those leads before taking
the square root. Missing-variable evaluations retain only the omitted
channel in this calculation.

\paragraph{Anomaly correlation coefficient.}
ACC measures the spatial agreement between forecast and observed anomalies
relative to the climatology of the verifying calendar day. Let
$X^{\mathrm{clim}}_{t+\ell,v}$ be the mean field for the corresponding
month and day over 1979--2018, expressed in standardized space. For each initialization $t$, lead $\ell$, and channel $v$, we compute
\begin{equation}
\begin{aligned}
a_{t,\ell,v}
&=\frac{\left\langle(\overline X_v-X_v^{\mathrm{clim}})
(X_v-X_v^{\mathrm{clim}})\right\rangle_w}
{\sqrt{\left\langle(\overline X_v-X_v^{\mathrm{clim}})^2\right\rangle_w
\left\langle(X_v-X_v^{\mathrm{clim}})^2\right\rangle_w}},\\
\operatorname{ACC}_{\ell}
&=\frac{1}{C|\mathcal I|}\sum_{v=1}^{C}\sum_{t\in\mathcal I}a_{t,\ell,v}.
\end{aligned}
\label{eq:d-acc}
\end{equation}
The first line suppresses the common valid-time subscript $t+\ell$;
all fields refer to that date and channel.
Anomalies are used directly without additional spatial centering.
An ACC of one indicates aligned anomaly patterns; zero indicates no
weighted spatial correlation.

\paragraph{Fair continuous ranked probability score.}
Fair-CRPS~\cite{ferro2014fairscores} evaluates the sampled predictive
distribution. For standardized fields, the distance used in
Sec.~\ref{app:c-crps} is
\begin{equation}
d_w(U,V)=\frac{1}{C}\sum_{v=1}^{C}\langle|U_v-V_v|\rangle_w.
\label{eq:d-distance}
\end{equation}
For $K>1$ independently sampled members, the score at one initialization is
\begin{equation}
\begin{aligned}
F_{t,\ell}={}&\frac{1}{K}\sum_{k=1}^{K}
d_w(\widehat X^{(k)}_{t+\ell},X_{t+\ell})\\
&-\frac{1}{K(K-1)}\sum_{k<k'}
d_w(\widehat X^{(k)}_{t+\ell},\widehat X^{(k')}_{t+\ell}).
\end{aligned}
\label{eq:d-fair-crps}
\end{equation}
The reported normalized fair-CRPS is
$|\mathcal I|^{-1}\sum_{t\in\mathcal I}F_{t,\ell}$.
The pairwise sum includes each unordered pair once; its normalization
corrects for finite ensemble size. Lower values indicate better
probabilistic forecasts. With $K=2$, this expression recovers the
fine-tuning objective in Eq.~\eqref{eq:c-crps}. For variable-wise plots
in physical units, we score each channel separately and multiply its
normalized fair-CRPS by $\sigma_v$.

\paragraph{Downstream event evaluation.}
Land and ocean event forecasts use the area-weighted critical success
index in Eq.~\eqref{eq:c-csi}, with thresholds, persistence rules, and
scoring windows specified in Sec.~\ref{app:c-downstream}. Continuous-field
RMSE in the downstream comparison aggregates member-wise squared errors
over the scored window before taking the square root; daily point-forecast
RMSE in Eq.~\eqref{eq:d-rmse} evaluates the ensemble mean.

\subsection{Supplementary Experimental Results}
\label{app:d-results}

\begin{figure*}[!t]
\centering
\includegraphics[width=\textwidth]{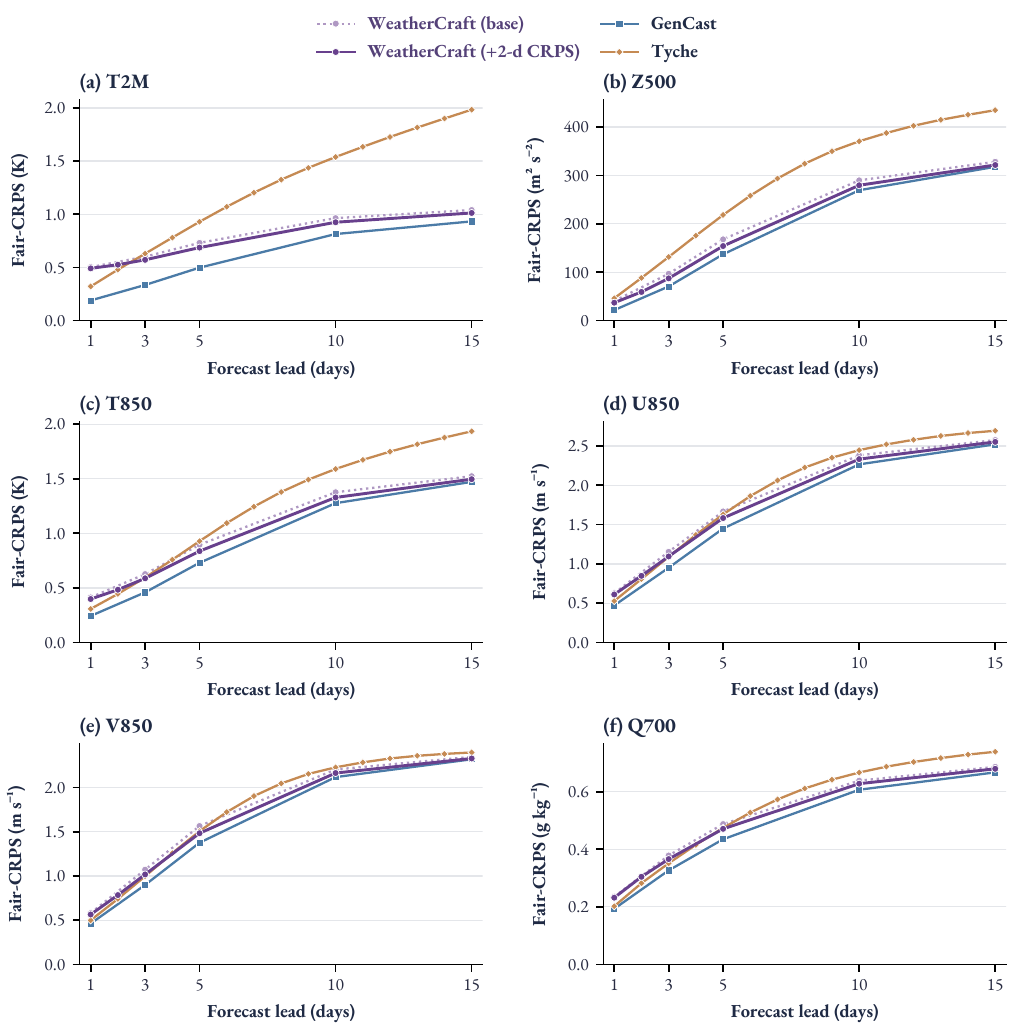}
\caption{\textbf{Variable-wise probabilistic forecasting.} Fair-CRPS of eight-member forecasts in the indicated physical units. ``+2-d CRPS'' denotes WeatherCraft after the two-stage fair-CRPS curriculum.}
\label{fig:d-fair-crps}
\end{figure*}

\paragraph{Variable-wise forecasting skill.}
Tables~\ref{tab:d-variable-rmse} and~\ref{tab:d-variable-acc} resolve daily
forecast skill into six representative surface and upper-air variables
at 5, 10, and 15 days. WeatherCraft achieves lower RMSE than every evaluated deterministic
baseline across all six variables at these leads. Figure~\ref{fig:d-fair-crps} compares the sampled
distributions for the same variables, showing WeatherCraft before and
after the two-stage fair-CRPS curriculum. The fine-tuned model improves 15-day fair-CRPS over
the pretrained model for all six variables. Relative to Tyche, it reduces
15-day T2M, Z500, and T850 fair-CRPS by 48.9\%, 26.1\%, and 22.7\%,
respectively.

\begin{table*}[!t]
\centering
\caption{\textbf{Variable-wise RMSE.} Units are K for T2M/T850, m$^2$\,s$^{-2}$ for Z500, m\,s$^{-1}$ for U850/V850, and g\,kg$^{-1}$ for Q700. \weathercraft{} uses the eight-member mean; deterministic baselines use one forecast. Bold and underlining mark the best and second-best values in each column and lead.}
\label{tab:d-variable-rmse}
\vspace{2pt}
\begingroup
\appfont
\setlength{\tabcolsep}{4pt}
\renewcommand{\arraystretch}{1.05}
\setlength{\aboverulesep}{1.6pt}
\setlength{\belowrulesep}{1.6pt}
\begin{tabularx}{\textwidth}{@{}p{0.20\textwidth}*{6}{>{\centering\arraybackslash}X}@{}}
\toprule[0.95pt]
\rowcolor{appheader}
\rule{0pt}{10.5pt}\textbf{Model} & \textbf{T2M} & \textbf{Z500} & \textbf{T850} & \textbf{U850} & \textbf{V850} & \textbf{Q700} \\
\midrule[0.55pt]
\appsubhead{7}{5-day forecasts}
\rowcolor{appstripe} ConvLSTM & 5.076 & 1091.5 & 4.361 & 7.328 & 7.696 & 1.805 \\
SimVP & 5.022 & 716.2 & 3.591 & 5.270 & 4.915 & 1.536 \\
\rowcolor{appstripe} SFNO & 2.697 & 692.5 & 3.107 & 5.084 & 4.782 & 1.434 \\
Pangu & 2.959 & 735.9 & 3.274 & 5.332 & 4.990 & 1.476 \\
\rowcolor{appstripe} OneForecast & \underline{1.997} & \underline{407.5} & \underline{2.073} & \underline{3.653} & \underline{3.468} & \underline{1.136} \\
FuXi & 2.330 & 607.3 & 2.739 & 4.582 & 4.424 & 1.245 \\
\rowcolor{appstripe} FengWu & 2.345 & 613.7 & 2.686 & 4.364 & 4.092 & 1.249 \\
GraphCast & 2.041 & 430.6 & 2.174 & 3.776 & 3.576 & 1.172 \\
\rowcolor{appstripe} STORM & 2.504 & 532.4 & 2.599 & 4.346 & 4.118 & 1.261 \\
\rowcolor{appours}
\rule{0pt}{10.8pt}\textcolor{apppurple}{\bfseries WeatherCraft} & \textbf{1.655} & \textbf{398.2} & \textbf{1.918} & \textbf{3.446} & \textbf{3.302} & \textbf{1.054} \\
\midrule[0.55pt]
\appsubhead{7}{10-day forecasts}
\rowcolor{appstripe} ConvLSTM & 10.407 & 1924.6 & 8.213 & 11.339 & 9.891 & 3.298 \\
SimVP & 5.933 & 1016.6 & 4.688 & 6.736 & 6.049 & 1.838 \\
\rowcolor{appstripe} SFNO & 3.470 & 958.9 & 4.005 & 6.100 & \underline{5.626} & 1.597 \\
Pangu & 3.928 & 1006.7 & 4.474 & 6.475 & 5.807 & 1.786 \\
\rowcolor{appstripe} OneForecast & \underline{3.188} & \underline{835.2} & \underline{3.687} & 5.998 & 5.690 & 1.656 \\
FuXi & 3.302 & 943.2 & 4.057 & 6.333 & 6.017 & 1.687 \\
\rowcolor{appstripe} FengWu & 3.360 & 910.2 & 3.819 & 6.100 & 5.699 & \underline{1.576} \\
GraphCast & 3.284 & 848.2 & 3.774 & \underline{5.974} & 5.666 & 1.647 \\
\rowcolor{appstripe} STORM & 3.673 & 934.6 & 4.007 & 6.338 & 5.936 & 1.719 \\
\rowcolor{appours}
\rule{0pt}{10.8pt}\textcolor{apppurple}{\bfseries WeatherCraft} & \textbf{2.330} & \textbf{704.0} & \textbf{3.010} & \textbf{4.935} & \textbf{4.715} & \textbf{1.352} \\
\midrule[0.55pt]
\appsubhead{7}{15-day forecasts}
\rowcolor{appstripe} ConvLSTM & 14.206 & 2568.3 & 11.344 & 13.538 & 11.135 & 3.382 \\
SimVP & 6.548 & 1139.0 & 5.238 & 7.288 & 6.333 & 2.018 \\
\rowcolor{appstripe} SFNO & 4.163 & 1063.7 & 4.494 & \underline{6.442} & \underline{5.810} & \underline{1.661} \\
Pangu & 4.825 & 1083.6 & 5.036 & 6.795 & 5.928 & 1.890 \\
\rowcolor{appstripe} OneForecast & 3.795 & 1018.8 & 4.364 & 6.856 & 6.421 & 1.885 \\
FuXi & \underline{3.673} & 1046.4 & 4.457 & 6.822 & 6.381 & 1.834 \\
\rowcolor{appstripe} FengWu & 4.088 & \underline{1007.9} & \underline{4.340} & 6.672 & 6.143 & 1.704 \\
GraphCast & 4.054 & 1019.4 & 4.512 & 6.748 & 6.307 & 1.848 \\
\rowcolor{appstripe} STORM & 4.398 & 1120.3 & 4.588 & 7.086 & 6.425 & 1.912 \\
\rowcolor{appours}
\rule{0pt}{10.8pt}\textcolor{apppurple}{\bfseries WeatherCraft} & \textbf{2.543} & \textbf{795.9} & \textbf{3.322} & \textbf{5.315} & \textbf{5.016} & \textbf{1.446} \\
\bottomrule[0.95pt]
\end{tabularx}
\endgroup
\end{table*}

\begin{table*}[!t]
\centering
\caption{\textbf{Variable-wise ACC.} \weathercraft{} uses the eight-member mean; deterministic baselines use one forecast. Bold and underlining mark the best and second-best values in each column and lead.}
\label{tab:d-variable-acc}
\vspace{2pt}
\begingroup
\appfont
\setlength{\tabcolsep}{4pt}
\renewcommand{\arraystretch}{1.05}
\setlength{\aboverulesep}{1.6pt}
\setlength{\belowrulesep}{1.6pt}
\begin{tabularx}{\textwidth}{@{}p{0.20\textwidth}*{6}{>{\centering\arraybackslash}X}@{}}
\toprule[0.95pt]
\rowcolor{appheader}
\rule{0pt}{10.5pt}\textbf{Model} & \textbf{T2M} & \textbf{Z500} & \textbf{T850} & \textbf{U850} & \textbf{V850} & \textbf{Q700} \\
\midrule[0.55pt]
\appsubhead{7}{5-day forecasts}
\rowcolor{appstripe} ConvLSTM & 0.072 & 0.042 & 0.100 & 0.106 & 0.027 & 0.146 \\
SimVP & 0.264 & 0.600 & 0.401 & 0.471 & 0.424 & 0.318 \\
\rowcolor{appstripe} SFNO & 0.390 & 0.578 & 0.475 & 0.419 & 0.385 & 0.324 \\
Pangu & 0.346 & 0.558 & 0.480 & 0.419 & 0.368 & 0.404 \\
\rowcolor{appstripe} OneForecast & \underline{0.703} & \underline{0.865} & \underline{0.792} & \underline{0.745} & \underline{0.725} & \underline{0.670} \\
FuXi & 0.579 & 0.703 & 0.637 & 0.594 & 0.542 & 0.589 \\
\rowcolor{appstripe} FengWu & 0.555 & 0.677 & 0.624 & 0.622 & 0.595 & 0.553 \\
GraphCast & 0.680 & 0.846 & 0.769 & 0.721 & 0.700 & 0.641 \\
\rowcolor{appstripe} STORM & 0.556 & 0.770 & 0.676 & 0.638 & 0.607 & 0.586 \\
\rowcolor{appours}
\rule{0pt}{10.8pt}\textcolor{apppurple}{\bfseries WeatherCraft} & \textbf{0.760} & \textbf{0.870} & \textbf{0.816} & \textbf{0.765} & \textbf{0.744} & \textbf{0.693} \\
\midrule[0.55pt]
\appsubhead{7}{10-day forecasts}
\rowcolor{appstripe} ConvLSTM & -0.049 & -0.099 & -0.036 & 0.000 & 0.013 & 0.001 \\
SimVP & 0.088 & 0.220 & 0.108 & 0.158 & 0.105 & 0.115 \\
\rowcolor{appstripe} SFNO & 0.130 & 0.181 & 0.134 & 0.148 & 0.096 & 0.139 \\
Pangu & 0.140 & 0.169 & 0.124 & 0.129 & 0.082 & 0.144 \\
\rowcolor{appstripe} OneForecast & \underline{0.338} & \underline{0.434} & \underline{0.357} & \underline{0.325} & \underline{0.264} & \underline{0.320} \\
FuXi & 0.233 & 0.281 & 0.228 & 0.223 & 0.146 & 0.256 \\
\rowcolor{appstripe} FengWu & 0.219 & 0.246 & 0.246 & 0.247 & 0.179 & 0.262 \\
GraphCast & 0.288 & 0.396 & 0.325 & 0.302 & 0.239 & 0.295 \\
\rowcolor{appstripe} STORM & 0.223 & 0.305 & 0.262 & 0.248 & 0.185 & 0.255 \\
\rowcolor{appours}
\rule{0pt}{10.8pt}\textcolor{apppurple}{\bfseries WeatherCraft} & \textbf{0.442} & \textbf{0.515} & \textbf{0.459} & \textbf{0.415} & \textbf{0.346} & \textbf{0.407} \\
\midrule[0.55pt]
\appsubhead{7}{15-day forecasts}
\rowcolor{appstripe} ConvLSTM & -0.034 & -0.088 & -0.020 & -0.000 & 0.011 & -0.012 \\
SimVP & 0.033 & 0.087 & 0.022 & 0.067 & 0.036 & 0.055 \\
\rowcolor{appstripe} SFNO & 0.009 & 0.059 & 0.012 & 0.081 & 0.044 & 0.086 \\
Pangu & 0.071 & 0.064 & 0.042 & 0.066 & 0.040 & 0.095 \\
\rowcolor{appstripe} OneForecast & \underline{0.164} & \underline{0.180} & \underline{0.133} & \underline{0.144} & \underline{0.090} & 0.154 \\
FuXi & 0.118 & 0.126 & 0.096 & 0.110 & 0.049 & 0.140 \\
\rowcolor{appstripe} FengWu & 0.101 & 0.096 & 0.075 & 0.114 & 0.059 & \underline{0.155} \\
GraphCast & 0.094 & 0.155 & 0.102 & 0.133 & 0.082 & 0.138 \\
\rowcolor{appstripe} STORM & 0.083 & 0.078 & 0.092 & 0.099 & 0.062 & 0.132 \\
\rowcolor{appours}
\rule{0pt}{10.8pt}\textcolor{apppurple}{\bfseries WeatherCraft} & \textbf{0.253} & \textbf{0.281} & \textbf{0.239} & \textbf{0.235} & \textbf{0.153} & \textbf{0.252} \\
\bottomrule[0.95pt]
\end{tabularx}
\endgroup
\end{table*}

\paragraph{Recovery from missing variables.}
We replace T2M, U10, V10, or MSLP with its calendar-day climatology in
all initial history fields and then perform the usual forecast rollout.
Table~\ref{tab:d-missing} reports the omitted channel's RMSE and ACC,
aggregated over leads 1--15. WeatherCraft maintains low missing-input
errors across all four variables, with particularly strong T2M and MSLP
forecasts.

\begin{figure*}[!t]
\centering
\includegraphics[width=0.90\textwidth]{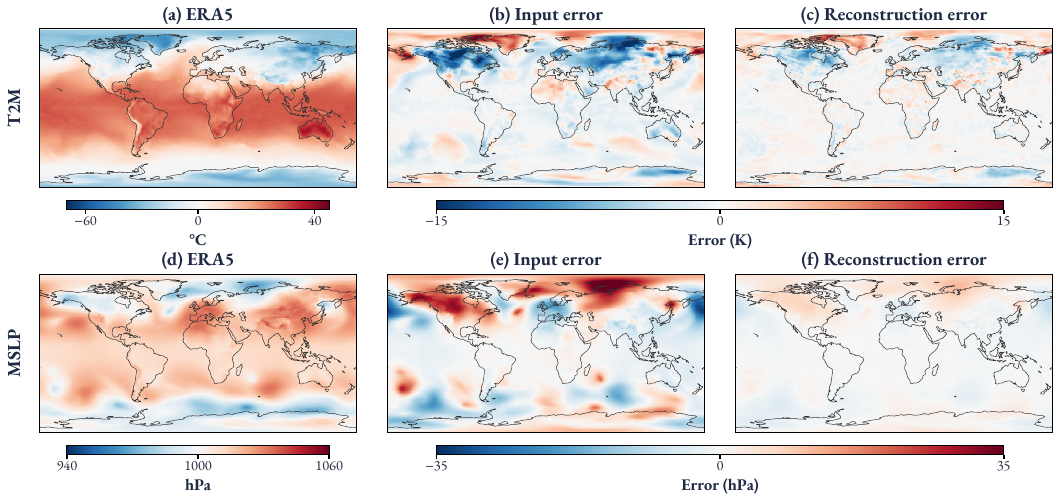}
\caption{\textbf{Recovery during tokenization.} ERA5 fields and errors before and after tokenizer reconstruction on 1 January 2020. T2M or MSLP is individually replaced by calendar-day climatology. Errors are relative to ERA5 and share a color scale within each row.}
\label{fig:d-recovery}
\end{figure*}

To locate where recovery begins, we pass the same incomplete histories
through the frozen tokenizer and decoder, before running any dynamics.
Table~\ref{tab:d-tokenizer-recovery} compares these reconstructions with
the climatology-filled inputs and clean-input reconstructions. Tokenizer
reconstruction reduces missing-channel RMSE by 44.6\% for T2M, 66.7\%
for U10, 65.1\% for V10, and 76.6\% for MSLP. Figure~\ref{fig:d-recovery}
shows the recovered spatial structure. These results establish that
substantial recovery occurs at the representation stage and support
the tokenizer's regularizing role under incomplete observations.

\begin{table*}[!t]
\centering
\caption{\textbf{Forecasts of missing input variables.} Scores evaluate only the omitted channel over leads $1$--$15$. Tyche, GenCast, and \weathercraft{} use eight-member means; the other models use a single forecast. Bold and underlining mark the best and second-best values in each column.}
\label{tab:d-missing}
\vspace{2pt}
\begingroup
\appfont
\setlength{\tabcolsep}{3pt}
\renewcommand{\arraystretch}{1.08}
\setlength{\aboverulesep}{1.8pt}
\setlength{\belowrulesep}{1.8pt}
\begin{tabularx}{\textwidth}{@{}p{0.18\textwidth}*{8}{>{\centering\arraybackslash}X}@{}}
\toprule[0.95pt]
\rowcolor{appheader}
\rule{0pt}{10.5pt}\textbf{Model} & \multicolumn{2}{c}{\textbf{T2M (K)}} & \multicolumn{2}{c}{\textbf{U10 (m\,s$^{-1}$)}} & \multicolumn{2}{c}{\textbf{V10 (m\,s$^{-1}$)}} & \multicolumn{2}{c@{}}{\textbf{MSLP (hPa)}} \\[-0.5pt]
\cmidrule(lr){2-3}\cmidrule(lr){4-5}\cmidrule(lr){6-7}\cmidrule(l){8-9}
 & \textbf{Clean} & \textbf{Missing} & \textbf{Clean} & \textbf{Missing} & \textbf{Clean} & \textbf{Missing} & \textbf{Clean} & \textbf{Missing} \\
\midrule[0.55pt]
\appsubhead{9}{RMSE (physical units) $\downarrow$}
\rowcolor{appstripe} SFNO & 3.2121 & 3.3698 & 3.6901 & 3.7314 & 3.6859 & 3.7354 & 7.2609 & 7.3686 \\
Pangu & 3.6005 & 3.6736 & 4.1101 & 4.2667 & 3.7873 & 3.9457 & 8.0505 & 9.0589 \\
\rowcolor{appstripe} FuXi & 2.9004 & 3.0934 & 3.7008 & 3.7075 & 3.7422 & 3.7603 & 7.0089 & 7.1207 \\
FengWu & 3.0326 & 3.1706 & 3.6908 & 3.7406 & 3.6740 & 3.7100 & 6.9119 & 7.2230 \\
\rowcolor{appstripe} OneForecast & 2.7911 & 3.1525 & 3.5207 & 3.5947 & 3.5511 & 3.5980 & 6.3902 & 7.1070 \\
Tyche & 2.4971 & 2.7939 & 2.8902 & 2.9284 & 2.9091 & 2.9511 & 5.7849 & \underline{5.9026} \\
\rowcolor{appstripe} GenCast & \textbf{1.8045} & \underline{2.1274} & \textbf{2.7331} & \textbf{2.8117} & \textbf{2.7826} & \textbf{2.8033} & \textbf{4.9883} & 6.5456 \\
\rowcolor{appours}
\rule{0pt}{10.8pt}\textcolor{apppurple}{\bfseries WeatherCraft} & \underline{2.0560} & \textbf{2.0547} & \underline{2.8408} & \underline{2.8506} & \underline{2.8980} & \underline{2.9195} & \underline{5.2088} & \textbf{5.1801} \\
\midrule[0.55pt]
\appsubhead{9}{ACC $\uparrow$}
\rowcolor{appstripe} SFNO & 0.2819 & 0.1357 & 0.3115 & 0.2955 & 0.2848 & 0.2708 & 0.3702 & 0.3553 \\
Pangu & 0.2858 & 0.1123 & 0.2560 & 0.1986 & 0.2557 & 0.2167 & 0.2949 & 0.0729 \\
\rowcolor{appstripe} FuXi & 0.4167 & 0.3112 & 0.4183 & 0.4114 & 0.3794 & 0.3706 & 0.4672 & 0.4469 \\
FengWu & 0.3937 & 0.2703 & 0.3654 & 0.3388 & 0.3302 & 0.3115 & 0.4373 & 0.3723 \\
\rowcolor{appstripe} OneForecast & 0.5004 & 0.3147 & 0.4982 & 0.4735 & 0.4691 & 0.4572 & 0.5703 & 0.4773 \\
Tyche & 0.4522 & 0.3329 & 0.5452 & 0.5340 & 0.5107 & 0.4997 & 0.5860 & \underline{0.5702} \\
\rowcolor{appstripe} GenCast & \textbf{0.6653} & \underline{0.5278} & \textbf{0.5993} & \textbf{0.5772} & \textbf{0.5639} & \textbf{0.5622} & \textbf{0.6667} & 0.4644 \\
\rowcolor{appours}
\rule{0pt}{10.8pt}\textcolor{apppurple}{\bfseries WeatherCraft} & \underline{0.5706} & \textbf{0.5675} & \underline{0.5676} & \underline{0.5637} & \underline{0.5300} & \underline{0.5205} & \underline{0.6384} & \textbf{0.6433} \\
\bottomrule[0.95pt]
\end{tabularx}
\endgroup
\end{table*}

\begin{table*}[!t]
\centering
\caption{\textbf{Recovery before dynamics prediction.} Target-variable reconstruction RMSE over both initial history days of $32$ test initializations. Only the indicated variable is replaced by climatology. Clean reconstruction uses complete inputs.}
\label{tab:d-tokenizer-recovery}
\vspace{2pt}
\begingroup
\appfont
\setlength{\tabcolsep}{4pt}
\renewcommand{\arraystretch}{1.12}
\setlength{\aboverulesep}{1.8pt}
\setlength{\belowrulesep}{1.8pt}
\begin{tabularx}{\textwidth}{@{}p{0.32\textwidth}*{4}{>{\centering\arraybackslash}X}@{}}
\toprule[0.95pt]
\rowcolor{appheader}
\rule{0pt}{10.5pt}\textbf{Input / reconstruction} & \textbf{T2M (K)} & \textbf{U10 (m\,s$^{-1}$)} & \textbf{V10 (m\,s$^{-1}$)} & \textbf{MSLP (hPa)} \\
\midrule[0.55pt]
\rowcolor{appstripe} Climatology input & 2.489 & 3.469 & 3.512 & 6.818 \\
Clean reconstruction & 0.933 & 0.389 & 0.346 & 0.433 \\
\rowcolor{appours}
\rule{0pt}{10.8pt}\textcolor{apppurple}{\bfseries Missing-input reconstruction} & \textbf{1.379} & \textbf{1.154} & \textbf{1.225} & \textbf{1.593} \\
\bottomrule[0.95pt]
\end{tabularx}
\endgroup
\end{table*}

\begin{figure*}[!t]
\centering
\includegraphics[width=0.90\textwidth]{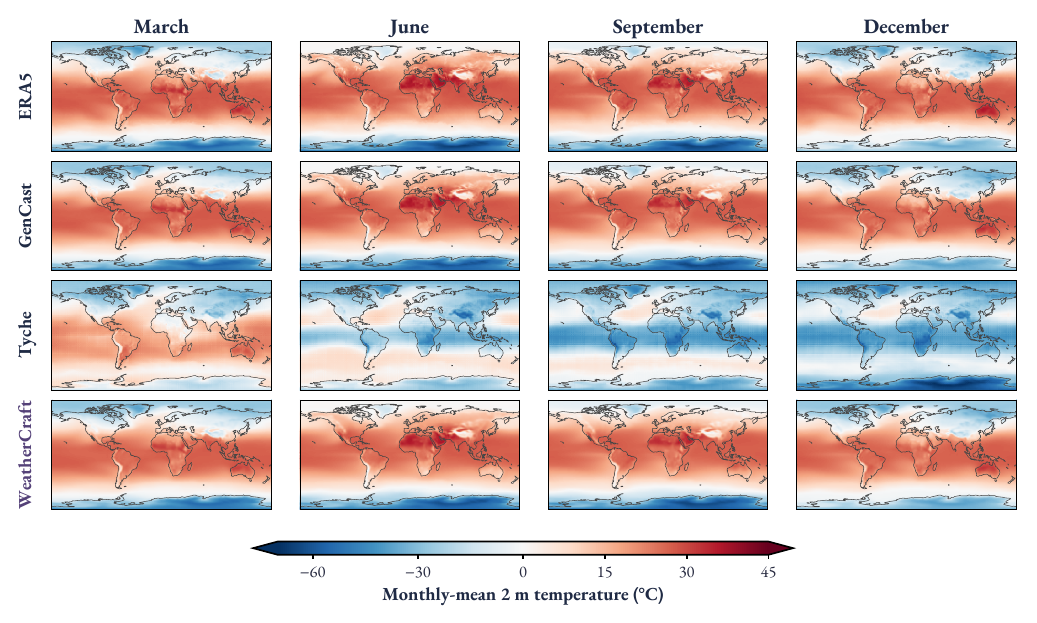}
\caption{\textbf{Monthly temperature fields during annual integration.} Monthly-mean T2M from the rollout initialized on 1 January 2019. Forecasts average eight members and all days within each indicated month; ERA5 shows the corresponding observed monthly means. All panels share a color scale.}
\label{fig:d-annual}
\end{figure*}

\paragraph{Seasonal spatial structure in year-long integration.}
We examine monthly-mean T2M fields from the 365-day rollout initialized
on 1 January 2019. Figure~\ref{fig:d-annual} compares March, June,
September, and December against ERA5 using eight-member ensemble means.
Monthly averaging reveals the spatial evolution of the seasonal state,
complementing the regional time series in the main text.
WeatherCraft maintains the tropical warm belt, polar cold regions, and
seasonal warming and cooling of continental interiors through December.
Latitude-weighted RMSE of its four monthly-mean T2M fields ranges from
1.41 to 2.03\,K. GenCast retains similar large-scale structure. Tyche
develops a broad cold drift over the tropics and continents. The monthly
patterns demonstrate coherent seasonal evolution during year-long
integration.

\subsection{Ablation Studies}
\label{app:d-ablation}

\paragraph{Finite scalar versus vector quantization.}
We replace FSQ with grouped VQ while retaining the encoder--decoder,
four parallel groups, and residual multiscale structure. Each group
uses codebooks of sizes 512 and 125, matching the factorized token
vocabulary. Table~\ref{tab:tokenizer_reconstruction} shows similar
reconstruction errors for the two quantizers. In forecasting,
FSQ achieves lower nRMSE at all reported leads and lower fair-CRPS
through day 10 (Table~\ref{tab:tokenizer_forecast}). At day 5, FSQ
reduces nRMSE from 0.3791 to 0.3721. Its stronger forecast skill at
these leads supports the use of a fixed scalar vocabulary for dynamics
modeling.

% Verified source values preserved; styled with EB Garamond matching main text.
\begin{table*}[!t]
\centering
\caption{\textbf{Forecasting with FSQ and VQ.} Eight-member test forecasts; nRMSE scores the ensemble mean and fCRPS denotes fair-CRPS. Bold and underlining mark the best and second-best values, ranked before rounding.}
\label{tab:tokenizer_forecast}
\vspace{2pt}
\begingroup
\appfont
\setlength{\tabcolsep}{2.5pt}
\renewcommand{\arraystretch}{1.12}
\setlength{\aboverulesep}{1.8pt}
\setlength{\belowrulesep}{1.8pt}
\begin{tabularx}{\textwidth}{@{}p{0.18\textwidth}*{10}{>{\centering\arraybackslash}X}@{}}
\toprule[0.95pt]
\rowcolor{appheader}
\rule{0pt}{10.5pt}\textbf{Tokenizer} & \multicolumn{2}{c}{\textbf{1-Day}} & \multicolumn{2}{c}{\textbf{3-Day}} & \multicolumn{2}{c}{\textbf{5-Day}} & \multicolumn{2}{c}{\textbf{10-Day}} & \multicolumn{2}{c@{}}{\textbf{15-Day}} \\[-0.5pt]
\cmidrule(lr){2-3}\cmidrule(lr){4-5}\cmidrule(lr){6-7}\cmidrule(lr){8-9}\cmidrule(l){10-11}
 & \textbf{nRMSE} & \textbf{fCRPS} & \textbf{nRMSE} & \textbf{fCRPS} & \textbf{nRMSE} & \textbf{fCRPS} & \textbf{nRMSE} & \textbf{fCRPS} & \textbf{nRMSE} & \textbf{fCRPS} \\
\midrule[0.55pt]
\rowcolor{appours}
\rule{0pt}{10.8pt}\textcolor{apppurple}{\bfseries Residual FSQ} & \textbf{0.1453} & \textbf{0.0602} & \textbf{0.2586} & \textbf{0.1065} & \textbf{0.3721} & \textbf{0.1526} & \textbf{0.5474} & \textbf{0.2252} & \textbf{0.5967} & \underline{0.2476} \\
Grouped VQ & \underline{0.1498} & \underline{0.0615} & \underline{0.2662} & \underline{0.1086} & \underline{0.3791} & \underline{0.1543} & \underline{0.5492} & \underline{0.2252} & \underline{0.5967} & \textbf{0.2471} \\
\bottomrule[0.95pt]
\end{tabularx}
\endgroup
\end{table*}

\paragraph{Residual versus independent multiscale tokenization.}
The independent variant quantizes the original encoder latent at each
scale and averages the upsampled codes. The residual variant instead
quantizes the error left by coarser scales and sums their corrections.
With the same encoder--decoder and scale grids, residual FSQ reduces
test reconstruction nRMSE from 0.0828 to 0.0664, a 19.9\% reduction
(Table~\ref{tab:tokenizer_reconstruction}). This comparison supports
allocating fine-scale tokens to information not already captured by
the coarse representation.

% Verified source values preserved; styled with EB Garamond matching main text.
\begin{table*}[t]
\centering
\caption{\textbf{Quantizer and multiscale reconstruction.} Latitude-weighted test nRMSE. Bold and underlining mark the best and second-best values.}
\label{tab:tokenizer_reconstruction}
\vspace{2pt}
\begingroup
\appfont
\setlength{\tabcolsep}{4pt}
\renewcommand{\arraystretch}{1.14}
\setlength{\aboverulesep}{1.8pt}
\setlength{\belowrulesep}{1.8pt}
\begin{tabularx}{\linewidth}{@{}X>{\centering\arraybackslash}p{0.32\linewidth}@{}}
\toprule[0.95pt]
\rowcolor{appheader}
\rule{0pt}{10.5pt}\textbf{Tokenizer} & \textbf{Test nRMSE $\downarrow$} \\
\midrule[0.55pt]
\rowcolor{appours}
\rule{0pt}{10.8pt}\textcolor{apppurple}{\bfseries Residual FSQ} & \underline{0.0664} \\
Grouped VQ & \textbf{0.0661} \\
\rowcolor{appstripe} Independent FSQ & 0.0828 \\
\bottomrule[0.95pt]
\end{tabularx}
\endgroup
\end{table*}

\paragraph{Parallel versus spatial-block generation within scales.}
Motivated by intra-scale dependency modeling in
Logit Refiner~\cite{li2026logitrefiner}, we test a spatial-block
autoregressive variant of the weather dynamics model. It retains the
frozen tokenizer and parameter count, orders $5\times10$ spatial blocks
in row-major order within each scale, and predicts tokens within a block
in parallel. Coarser scales precede finer ones.
Table~\ref{tab:spatial_block} compares this variant with the matched
scale-parallel model. Scale-parallel generation achieves lower nRMSE
and fair-CRPS at both evaluated leads; at day 5, nRMSE is 0.3707 versus
0.3791 and fair-CRPS is 0.1519 versus 0.1556. These results favor
parallel generation within each scale for the evaluated weather model.

% Verified source values preserved; styled with EB Garamond matching main text.
\begin{table*}[t]
\centering
\caption{\textbf{Within-scale generation order.} Eight-member forecasts on the same $32$ validation initializations. fCRPS denotes fair-CRPS.}
\label{tab:spatial_block}
\vspace{2pt}
\begingroup
\appfont
\setlength{\tabcolsep}{3pt}
\renewcommand{\arraystretch}{1.14}
\setlength{\aboverulesep}{1.8pt}
\setlength{\belowrulesep}{1.8pt}
\begin{tabularx}{\linewidth}{@{}p{0.36\linewidth}*{4}{>{\centering\arraybackslash}X}@{}}
\toprule[0.95pt]
\rowcolor{appheader}
\rule{0pt}{10.5pt}\textbf{Model} & \multicolumn{2}{c}{\textbf{3-Day}} & \multicolumn{2}{c@{}}{\textbf{5-Day}} \\[-0.5pt]
\cmidrule(lr){2-3}\cmidrule(l){4-5}
 & \textbf{nRMSE} & \textbf{fCRPS} & \textbf{nRMSE} & \textbf{fCRPS} \\
\midrule[0.55pt]
\rowcolor{appours}
\rule{0pt}{10.8pt}\textcolor{apppurple}{\bfseries Base (parallel)} & \textbf{0.2616} & \textbf{0.1074} & \textbf{0.3707} & \textbf{0.1519} \\
Spatial-block AR & \underline{0.2641} & \underline{0.1085} & \underline{0.3791} & \underline{0.1556} \\
\bottomrule[0.95pt]
\end{tabularx}
\endgroup
\end{table*}

\end{document}